\documentclass[preprint,10pt]{elsarticle}

\usepackage{amssymb}
\usepackage{amsmath}

\usepackage{amsmath,amssymb,graphicx}
\usepackage{epstopdf}
\usepackage{amsfonts,bm,booktabs,multirow,color}
\usepackage{threeparttable}
\usepackage{subfigure}
\usepackage{url}
\usepackage[noend]{algpseudocode}
\usepackage{algorithmicx,algorithm}
\usepackage{makecell}

\journal{Journal}

\begin{document}

\begin{frontmatter}



\title{MSS-PAE: Saving Autoencoder-based Outlier Detection from Unexpected Reconstruction}


\author[1]{Xu Tan}
\ead{xutan@ieee.org}

\author[2]{Jiawei Yang}
\ead{jiaweiyang@ieee.org}

\author[1]{Junqi Chen}
\ead{jqchen@ieee.org}

\author[3]{Sylwan Rahardja}
\ead{sylwanrahardja@ieee.org}

\author[1,4]{Susanto Rahardja\corref{mycorrespondingauthor}}
\ead{susantorahardja@ieee.org}

\cortext[mycorrespondingauthor]{Corresponding author}

\affiliation[1]{organization={School of Marine Science and Technology, Northwestern Polytechnical University},
            city={Xi'an},
            country={P.R.China}}
\affiliation[2]{organization={Faculty of Technology, University of Turku},
	city={Turku},
	country={Finland}}
\affiliation[3]{organization={School of Computing, University of Eastern Finland},
	city={Joensuu},
	country={Finland}}
\affiliation[4]{organization={Engineering Cluster, Singapore Institute of Technology},
	country={Singapore}}

\begin{abstract}
AutoEncoders (AEs) are commonly used for machine learning tasks due to their intrinsic learning ability. This unique characteristic can be capitalized for Outlier Detection (OD). However conventional AE-based methods face the issue of overconfident decisions and unexpected reconstruction results of outliers, limiting their performance in OD. To mitigate these issues, the Mean Squared Error (MSE) and Negative Logarithmic Likelihood (NLL) were firstly analyzed, and the importance of incorporating aleatoric uncertainty to AE-based OD was elucidated. Then the Weighted Negative Logarithmic Likelihood (WNLL) was proposed to adjust for the effect of uncertainty for different OD scenarios. Moreover, the Mean-Shift Scoring (MSS) method was proposed to utilize the local relationship of data to reduce the issue of false inliers caused by AE. Experiments on 32 real-world OD datasets proved the effectiveness of the proposed methods. The combination of WNLL and MSS achieved 41\% relative performance improvement compared to the best baseline. In addition, MSS improved the detection performance of multiple AE-based outlier detectors by an average of 20\%. The proposed methods have the potential to advance AE's development in OD.
\end{abstract}



\begin{keyword}


Outlier detection \sep Anomaly detection \sep Autoencoder \sep Uncertainty estimation \sep Mean-shift

\end{keyword}

\end{frontmatter}



\section{Introduction}
\label{sec:introduction}
Outlier detection (OD), also known as anomaly detection, is a fundamental technique in the field of data mining. An outlier can be defined as ``one that appears to deviate markedly from other members of the sample in which it occurs'' \cite{grubbs1969procedures}. Outliers exhibit two main properties: (1) different from inliers with respect to their features; (2) rare in a dataset compared to inliers \cite{goldstein2016comparative}. The objective of OD is to identify the outliers from the normal samples (inliers), for the purpose of keeping the data clean and safe, or identifying the abnormal situations and behaviors. 
OD has many applications, such as fraud detection \cite{yang2022click}, network intrusion detection \cite{ghosh2019resisting}, disease diagnosis \cite{yang2020classification}, video surveillance \cite{li2019video} and trajectory analysis \cite{yang2022mipo}. 
Generally, OD was treated as an unsupervised or semi-supervised task, since the ground-truth labels were hard to acquire in real-world applications \cite{domingues2018comparative}---which means most of the data consists of unidentified inliers and outliers.

Numerous OD methods had been proposed over the years. They could be categorized into different classes. Proximity-based methods \cite{ramaswamy2000efficient, breunig2000lof, du2016novel, yang2021mean}  measured the similarities between inliers and outliers in the original data space. Statistical-based methods \cite{tang2015outlier, dalatu2017comparative, latecki2007outlier} modeled the statistic distribution of inliers. Classification-based methods \cite{scholkopf2001estimating, tax2004support, ruff2019deep} transformed the data to the substitute space, then identified outliers by dividing the substitute space. Ensemble-based methods \cite{liu2008isolation, tan2022sparse, pevny2016loda, zhao2021suod} combined several weak outlier detectors to produce a strong detector by utilizing the stochasticity and diversity of data. Probabilistic-based methods \cite{goldstein2012histogram, li2022ecod} analyzed the probabilistic distribution of the attributes of the data. 

\textcolor{black}{Traditional OD methods usually countered problems when facing high data dimensionality and complex data manifolds, due to their weak feature extraction ability. Thus in recent years, deep learning techniques with strong learning abilities piqued the curiosity of experts in OD.}
Learning-based methods, such as autoencoder-based (AE-based) \cite{hawkins2002outlier, zhou2017anomaly, an2015variational, lai2020robust, guo2022unsupervised, angiulli2022mathrm} and generative-adversarial-network-based (GAN-based) \cite{schlegl2017unsupervised, akcay2018ganomaly, ibrahim2020vae, yang2022memory}, utilized the strong learning ability of the neural network to capture the inherent feature of the inliers. AE-based methods assumed that the inliers could be reconstructed better than the outliers. GAN-based methods created outliers through the generator, or distinguished between inliers and outliers through the discriminator. At present, GAN-based methods are generally hard to train because of their adversarial property, thus AE-based methods are the most popular among all learning-based methods \cite{chalapathy2019deep}. 

\textcolor{black}{Most AE-based methods utilize the reconstruction error, such as the Mean Squared Error (MSE), to detect outliers. Optimizing MSE can be interpreted as a special case of maximizing likelihood with an underlying Gaussian error model \cite{nix1994estimating}. It assumes the variance term of the logarithmic likelihood function to be 1, independent of the data instances or attributes. Although this assumption simplifies the optimization and architecture of AE, it limits AE's discriminating power. More concretely, AE with MSE loss may make overconfident yet wrong decisions, which is harmful to OD tasks.}

\textcolor{black}{Some researchers had considered the aforementioned variance term (also known as the aleatoric uncertainty) to achieve better OD performance \cite{an2015variational, pol2019anomaly, mao2020abnormality, choi2018waic, yong2022bayesian}. Nevertheless, they lack the analysis of its effect on the OD task, especially for OD on tabular data. Hence, this work elucidates the mechanism of how the variance term impacts AE's reconstruction result, and explains the reasons for the performance improvement that it brings to OD. Upon further analysis of the trade-off within the Negative Logarithmic Likelihood (NLL) function, it was discovered that a proper weighting value can significantly improve OD performance, which was neglected in the literature. Consequently, the \textit{Weighted Negative Logarithmic Likelihood (WNLL)}, a score function effective and flexible for various OD tasks, was proposed. 
}

Secondly, conventional AE-based OD methods suffered from the abnormal reconstruction problem, where some outliers are unexpectedly reconstructed well, making them difficult to distinguish from the inliers \cite{merrill2020modified}. This phenomenon can be explained by the absence of conventional AEs in considering local relationship, instead only capturing global features of data. 
In this paper, the issue is mitigated by modifying the scoring strategy of AE, producing the \textit{Mean-Shift outlier Scoring (MSS)} method. This method employed the mean-shifted data point instead of the original data point to calculate the outlier scores. Thus, when an outlier is reconstructed with a relatively high likelihood, the reconstruction result will not resemble the mean-shifted result of the input, leading to a high outlier score. This method is convenient and effective, and can be easily applied to most AE-based methods, which will be shown in Section \ref{sec:exp}. 

\begin{figure}[t]
	\centering
	\centerline{\includegraphics[width=8cm]{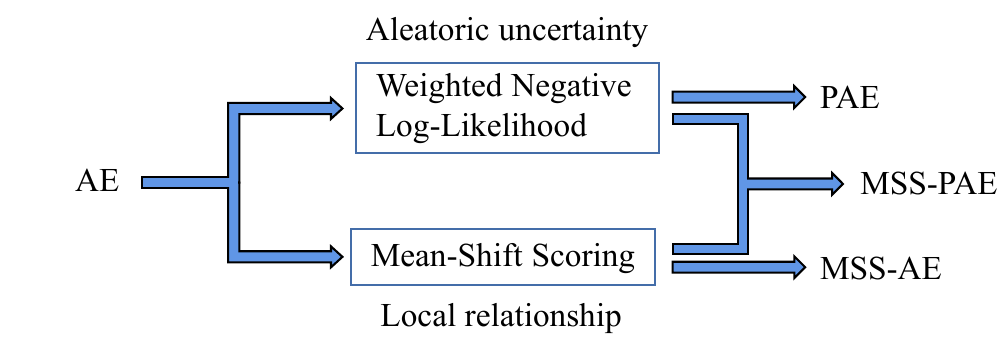}}
	
	\caption{\textcolor{black}{An illustration of how this work will improve AE-based OD methods. WNLL incorporates the aleatoric uncertainty to alleviate the overconfidence issue of AE, and MSS introduces the local relationship to detect well-reconstructed outliers.}
	}
	\label{fig:demo}
\end{figure}

\textcolor{black}{An illustration of how this work will improve AE-based OD methods is shown in Fig. \ref{fig:demo}. WNLL factors in the aleatoric uncertainty to alleviate the overconfidence issue of AE, and MSS introduces the local relationship to detect well-reconstructed outliers.}

In summary, the main contributions of this work are as follows:
\begin{itemize}
	\item Theoretically analyzing the overconfidence issue in conventional AE, and interpreting the effect of aleatoric uncertainty for the reconstruction task and AE-based OD.
	\item Proposing WNLL, which mitigates the overconfidence issue for AE-based OD, and significantly improves the detection performance by balancing the effect of aleatoric uncertainty.
	\item Proposing MSS, which introduces local relationships to reduce the harm of unexpected well-reconstructed outliers in AE, thus improving the robustness and accuracy. In addition to its efficacy, it can be easily applied to other AE-based OD methods. 
	\item Conducting experiments on 32 real-world OD datasets, comparing the proposed methods with 5 typical AE-based OD methods and 8 non-AE-based state-of-the-art (SOTA) OD methods. The experimental results proved the effectiveness and superiority of our methods.
\end{itemize}

\textcolor{black}{The rest of the paper is organized as follows: In Section \ref{sec:rel}, the principle of AE-based OD, typical AE-based OD methods proposed in recent years, and the studies quantifying uncertainty in neural networks are reviewed. 
	In Section \ref{sec:met}, the proposed NLL, WNLL, and MSS methods are introduced in detail, and their theoretical advantages are analyzed. Then the experimental settings are described and the results are reported in Section \ref{sec:exp}. 
	Finally, a conclusion is drawn in Section \ref{sec:con}.}

\section{Preliminary and Related Works}
\label{sec:rel}

\subsection{Autoencoder for outlier detection}
An AE is a neural network that is trained to reproduce its input to its output \cite{goodfellow2016deep}. It comprises of two main components: the encoder and the decoder. Considering a traditional fully-connected AE, given a multivariate data vector $\mathbf{x} \in \mathbb{R}^{D}$ as the input, the encoder transforms it to a lower-dimensional latent representation, which retains the most critical features of the input data. The latent representation $\mathbf{z}$ output by the encoder is given by 
\begin{equation}
	\label{eq1}
	\mathbf{z} = {\rm Encoder}(\mathbf{x}, \theta),
\end{equation}
where $\theta$ denotes the network parameters of the encoder. Then, the decoder reconstructs the input using the latent representation and the output $\hat{\mathbf{x}}$ is expressed as
\begin{equation}
	\label{eq2}
	\hat{\mathbf{x}} = {\rm Decoder}(\mathbf{z}, \phi),
\end{equation}
where $\phi$ denotes the network parameters of the decoder.

Since the fully-connected layer can be viewed as the linear dimensionality reduction transformation, the effect of the encoder is similar to the principle components analysis, which aims to find the main components reflecting the features of the original data \cite{wold1987principal}. However, complicated real-world data are difficult to analyze linearly, so a nonlinear activation function is applied after each fully-connected layer to increase the expressing ability of the network. Common typical activation functions are Sigmoid, Tanh, Rectified Linear Unit (ReLU), and ReLU's variants.

The optimization and outlier score computation of an AE both relies on its loss function. Since the target of the AE is to reconstruct the input data as the output, the bias between the input and output can be used as the loss function. This is also known as the reconstruction error. The most popular loss function of the conventional AE is the Mean Squared Error function, which is established by the following equation:
\begin{equation}
	\label{eq:mse}
	{\rm MSE}(\mathbf{x}, \hat{\mathbf{x}}) = ||\mathbf{x} - \hat{\mathbf{x}}||_{2}^{2},
\end{equation}
where $|| \cdot ||_{2}$ denotes the L2-norm. 

AE first transforms the input $\mathbf{x}$ to the latent $\mathbf{z}$, then transforms $\mathbf{z}$ to the output $\hat{\mathbf{x}}$. The dimension of the latent presentation $\mathbf{z}$ is designed to be smaller than the input layer. It served as a bottleneck, allowing only critical information to pass through \cite{thompson2002implicit}. Thus, the AE is forced to capture the key features of the data to better reconstruct the input. 

Considering a dataset that contains both inliers and outliers. Inliers are abundant and generally showed similar patterns, thus their features can easily be learned by an AE. Outliers are few and different in the dataset, hence their features are difficult for an AE to learn. As a consequence, inliers tend to be reconstructed with lower error margins, but outliers tend to have higher reconstruction errors.

\subsection{Autoencoder-based outlier detection methods}
\label{ssec:tae}
\textcolor{black}{AE plays a crucial role in unsupervised learning-based OD methods. AE can either be used separately or combined with other approaches \cite{zong2018deep, angiulli2022mathrm}. In addition to applying AE in various OD applications, some researchers have focused on enhancing the performance of AE itself to better align with the requirements of OD.
	Zhou \textit{et al.} proposed the Robust Deep Autoencoders (RDA) \cite{zhou2017anomaly}, motivated by the robust principal component analysis. RDA proposed a new loss function that can iteratively remove anomalous components from the training set, reducing AE contamination. 
	Chen \textit{et al.} proposed the Randomized Neural Network for Outlier Detection (RandNet) \cite{chen2017outlier} to improve the robustness of AE by training multiple AEs with random architectures on different subsets of training data. 
	An \textit{et al.} proposed a variational-autoencoder-based (VAE-based) OD method \cite{an2015variational}. It used the variational inference and reconstruction probability to get more principled and objective outlier scores. 
	Ishii \textit{et al.} proposed the Low-Cost Autoencoder (LCAE) \cite{ishii2019low}. It only used data with low reconstruction error in each training epoch, reducing the contamination of the outliers during training. 
	Gong \textit{et al.} proposed a Memory-augmented Autoencoder (MemAE) \cite{gong2019memorizing}, which improved the robustness of the AE by reconstructing samples from a limited number of recorded representative normal patterns. 
	Lai \textit{et al.} proposed an AE model based on the Robust Subspace Recovery layer (RSRAE) \cite{lai2020robust}. It used the robust subspace recovery layer to increase the reconstruction difficulty of outliers. 
	Similarly, Yu \textit{et al.} proposed an AE model based on the orthogonal projection constraints (OPCAE)\cite{yu2021autoencoder}, which used the Kautlr-Thomas transformation to preserve only inliers information after encoding. 
	Guo \textit{et al.} proposed an AE architecture called Feature Decomposition Autoencoder (FDAE) \cite{guo2022unsupervised}, which integrated the benefits from RDA \cite{zhou2017anomaly} and RSRAE \cite{lai2020robust}.
}

\textcolor{black}{In summary, most of the recent AE-based methods were dedicated to protect AE against the detrimental effects of outliers during training. However, they mostly have complicated network architectures and training procedures. This makes them difficult to use and thus achieve less satisfactory performance in practice. Additionally, they did not realize that AE sometimes make overconfident yet wrong decisions in unintended regions, which will be discussed in detail in the next subsection.}

\subsection{\textcolor{black}{Quantify uncertainty in neural networks}}
\label{ssec:qu}
\textcolor{black}{Despite neural networks' impressive capability in various machine learning tasks, they tend to produce overconfident decisions in some cases \cite{lakshminarayanan2017simple, nalisnick2018deep, hafner2020noise, yong2022bayesian, sensoy2020uncertainty}. For example, a regression model may output confident predictions on the regions without any training data \cite{hafner2020noise}. The overconfident yet wrong decisions can cause dire consequences in practical applications, especially for high-risk applications, such as autonomous driving and medical diagnosis \cite{zhang2023survey}. While superior models and extensive high-quality training data can reduce wrong decisions, it remains impractical to account for every possible situation \cite{staahl2020evaluation}. Therefore, it is crucial for a neural network to \textit{know what it knows} \cite{lakshminarayanan2017simple}, that is, the uncertainty quantification.}

\textcolor{black}{Conventionally, the uncertainty of neural networks can be divided into two main types: aleatoric uncertainty and epistemic uncertainty \cite{der2009aleatory, kendall2017uncertainties, sensoy2020uncertainty}. Aleatoric uncertainty, also known as data uncertainty, represents the noise or randomness inherent in observational data. It can be further categorized into heteroscedastic and homoscedastic uncertainty, depending on whether the noise is data instance dependent \cite{kendall2017uncertainties, staahl2020evaluation}. Epistemic uncertainty, also known as model uncertainty, represents the network's ignorance about whether model parameters captured the underlying regularity of data. It can be caused by erroneous training, weak model, or lack of knowledge due to unknown samples or insufficient training data \cite{gawlikowski2023survey}. Due to the different properties of aleatoric uncertainty and epistemic uncertainty, researchers employed different approaches to quantify them. For heteroscedastic aleatoric uncertainty, the common way is to let networks estimate the variance of the observation noise \cite{kendall2017uncertainties, staahl2020evaluation}. To quantify epistemic uncertainty, Bayesian neural networks \cite{blundell2015weight, gal2016dropout, nemeth2021stochastic} or ensemble techniques \cite{lakshminarayanan2017simple, pearce2020uncertainty} are commonly utilized.}

\textcolor{black}{Since uncertainty is strongly associated with the data distribution, researchers realized that quantifying uncertainty is beneficial to AE-based outlier detection. Some researches quantified aleatoric uncertainty. An \textit{et al.} proposed to use the reconstruction probability instead of MSE as the loss function of VAE, and highlighted that anomalous data will have greater aleatoric uncertainty \cite{an2015variational}. Pol \textit{et al.} employed NLL to train VAE and aleatoric-uncertainty-normalized reconstruction error as the outlier score \cite{pol2019anomaly}. Mao \textit{et al.} utilized the NLL-trained AE to detect abnormal pixels, for which the reconstruction error to aleatoric uncertainty ratio is high. Some researchers quantified epistemic uncertainty instead. Legrand \textit{et al.} and Baur \textit{et al.} trained Bayesian AE with Monte-Carlo Dropout to compute epistemic uncertainty \cite{legrand2019use, baur2020bayesian}. Legrand \textit{et al.} combined epistemic uncertainty with the reconstruction error as the final outlier score, while Baur \textit{et al.} pointed out that lesional tissues on brain MR images showed lower epistemic uncertainty. Daxberger \textit{et al.} trained a Bayesian VAE with Markov chain Monte Carlo to compute epistemic uncertainty, and treated data with large epistemic uncertainty as the outliers \cite{daxberger2019bayesian}. Park \textit{et al.} trained only the encoder of Bayesian VAE, and computed an integrated outlier score function that encompassed both Kullback-Leibler divergence and epistemic uncertainty, utilizing Monte-Carlo Dropout \cite{park2022interpreting}. Some researchers also explored the combination of aleatoric uncertainty and epistemic uncertainty \cite{choi2018waic, yong2021bayesian}}

\textcolor{black}{Although the uncertainty quantification had been applied to previous studies, the intricacies of its effect on the OD task had not been elucidated, especially for OD on general tabular data. It is worth noting that, the implication of uncertainty in various machine learning or OD scenarios may vary. For example, in regression tasks, aleatoric uncertainty reflects the noise in the input and target measurements that networks are unable to learn to correct. But in classification tasks, it reflects the deficiency of information to identify one class of data \cite{gawlikowski2023survey}. Conventional uncertainty studies largely focus on regression, classification, or segmentation. However, the situation for OD tasks with reconstruction models such as AE is different. In this paper, we detailedly analyze the aleatoric uncertainty in AE, and explain its advantages for tabular OD.}




\section{Saving AE from unexpected reconstruction}
\label{sec:met}


In this section, we offered insights for the two reasons of why conventional AE suffers from the unexpected reconstruction problem that lead to unsatisfied outlier detection performance in some specific cases. Moreover, we proposed two ways to address the issue effectively.

\subsection{AE is overconfident about its reconstruction result}
\label{ssec:apre}

The optimization target of AE can be interpreted as maximizing the reconstruction likelihood with certain underlying probability distribution according to the data noise or randomness.
Assuming the distribution is the diagonal multivariate Gaussian distribution, Eq. \ref{eq4} is established:
\begin{equation}
	\label{eq4}
	\left\{
	\begin{aligned}
		& \mathbf{x}^{\rm rec} \sim  {\rm N}(\bm{\mu},\bm{\Sigma}), \\
		& \bm{\mu} =  [\mu_{1},\mu_{2},\cdots,\mu_{D}]^{T}, \\
		& \bm{\sigma}^{2} =  [\sigma_{1}^{2},\sigma_{2}^{2},\cdots,\sigma_{D}^{2}]^{T}, \\
		& \bm{\Sigma} =  {\rm diag}(\bm{\sigma}^{2}),
	\end{aligned}
	\right.
\end{equation}
where $\mathbf{x}^{\rm rec}$ denotes the reconstruction random variable of input $\mathbf{x}$; ${\rm diag}(\cdot)$ denotes the diagonal matrix; $\bm{\mu}$ denotes the reconstruction mean of AE
; $\bm{\sigma}^{2}$ denotes the reconstruction variance
. This gives rise to Eq. \ref{eq5}:
\begin{equation}
	\label{eq5}
	P(\mathbf{x}^{\rm rec}) = \frac{1}{(2\pi)^{D/2}}\frac{1}{|\bm{\Sigma}|^{1/2}}e^{-\frac{1}{2}(\mathbf{x}^{\rm rec}-\bm{\mu})^{T}\bm{\Sigma}^{-1}(\mathbf{x}^{\rm rec}-\bm{\mu})},
\end{equation}
where $P(\cdot)$ denotes the probability density function, $|\bm{\Sigma}|$ denotes the determinant of the matrix $\bm{\Sigma}$, and $\bm{\Sigma}^{-1}$ denotes the inverse matrix of $\bm{\Sigma}$. The logarithmic probability density function may be expressed as Eq. \ref{eq6}:
\begin{equation}
	\label{eq6}
	\ln P(\mathbf{x}^{\rm rec}) = -\frac{D}{2}\ln 2\pi -\frac{1}{2}\sum_{d=1}^{D}{\ln \sigma_{d}^{2}} -\frac{1}{2}\sum_{d=1}^{D}{\frac{(x_{d}^{\rm rec}-\mu_{d})^{2}}{\sigma_{d}^{2}}}.
\end{equation}

If all variance values are assumed to be 1, then 
\begin{equation}
	\label{eq7}
	\begin{aligned}
		\sigma_{1}^{2} = \sigma_{2}^{2} = \cdots &= \sigma_{D}^{2} = 1, \\
		\bm{\Sigma} = &\mathbf{I},
	\end{aligned}
\end{equation}
where $\mathbf{I}$ denotes the identity matrix. Therefore, the logarithmic probability density function, also known as the log-likelihood of $\mathbf{x}^{\rm rec}$ can be expressed by Eq. \ref{eq8}:
\begin{equation}
	\label{eq8}
	\begin{aligned}
		\ln P(\mathbf{x}^{\rm rec}) &= -\frac{D}{2}\ln 2\pi - \frac{1}{2}\sum_{d=1}^{D}{(x_{d}^{\rm rec}-\mu_{d})^{2}} \\
		&= C - \frac{1}{2}||\mathbf{x}^{\rm rec} - \bm{\mu}||_{2}^{2},
	\end{aligned}
\end{equation}
where $C$ denotes a constant which will not affect the optimization process.

Conventional AEs aim to reconstruct the input $\mathbf{x}$, which implies maximizing the log-likelihood $\ln P(\mathbf{x}^{\rm rec})$ when $\mathbf{x}^{\rm rec} = \mathbf{x}$, and the mean value $\bm{\mu}$ is treated as the output $\hat{\mathbf{x}}$ of AE. Therefore, the optimization objective is to maximize $C - \frac{1}{2}||\mathbf{x} - \hat{\mathbf{x}}||_{2}^{2}$. Evidently, it is equivalent to minimizing $||\mathbf{x} - \hat{\mathbf{x}}||_{2}^{2}$, which is exactly the form of general MSE. So given the training set $\mathbf{X} \in \mathbb{R}^{N \times D}$, the loss function of the conventional AE can be represented by Eq. \ref{eq9}:
\begin{equation}
	\label{eq9}
	{\rm MSE}(\mathbf{X}, \hat{\mathbf{X}}) = \frac{1}{N}\sum_{i=1}^{N}{||\mathbf{x}_{i} - \hat{\mathbf{x}_{i}}||_{2}^{2}},
\end{equation}
where $N$ denotes the number of samples in the training set.

\begin{figure}[t]
	\centering
	\scalebox{0.8}{AE \hspace{4.3cm} PAE \hspace{0.7cm}} \\
	\vspace{0.2cm}
	\centerline{\includegraphics[width=7.8cm]{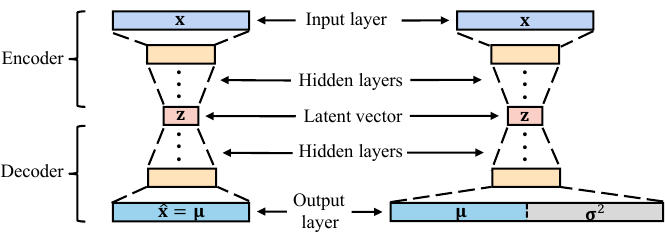} \hspace{1cm}}
	
	\caption{The typical structure of the AE and PAE. AE has a symmetrical structure and the size of the latent vector is much smaller than the input layer, while PAE has an output layer twice the size of its input layer.
	}
	\label{fig:ae}
\end{figure}

Experts on AE predominantly used the form of Eq. \ref{eq8} or its variants, such as $|| \cdot ||_{2}$ (root mean squared error) or $|| \cdot ||_{1}$ (mean absolute error) as the loss function. However, the usage of the MSE loss is under the assumption that the variance values $\bm{\sigma}^{2}$ of all the reconstruction probability distributions are 1. With this assumption, AE only estimates the mean values $\bm{\mu}$ of the distribution.
\textcolor{black}{The assumption simplifies the optimization and architecture of AE, but it is a sub-optimal solution since it does not reflect the practical situation. It assumes that the randomness of each data instance and attribute is constant and independent (i.e., homoscedastic), which is extremely rare in reality. This deficiency prevents AE with MSE from achieving higher reconstruction accuracy, and will raise the overconfidence issue for OD tasks. }
\textcolor{black}{Consequently, it is necessary to factor in the variance $\bm{\sigma}^{2}$, which leads to the NLL loss:}
\begin{equation}
\label{eq10}
{\rm NLL}(\mathbf{X}, \bm{\mu}, \bm{\sigma}) = \frac{1}{N}\sum_{i=1}^{N}{\sum_{d=1}^{D}{\frac{(x_{i,d}-\mu_{i,d})^{2}}{\sigma_{i,d}^{2}}} + \ln \sigma_{i,d}^{2}}.
\end{equation}
where $x_{i,d}$, $\mu_{i,d}$ and $\sigma_{i,d}^{2}$ denote the $d$-th attribute of the sample $\mathbf{x}_{i}$, output mean values $\bm{\mu}_{i}$, and output variance values $\bm{\sigma}_{i}^{2}$, respectively. \textcolor{black}{It is a simple deformation of Eq. \ref{eq6}, after elimination of the constant term, coefficients, and negation.
Since the computation of NLL required two variables, $\bm{\mu}_{i}$ and $\bm{\sigma}_{i}^{2}$, it required a network that has two outputs correspondingly. Therefore, the conventional AE was modified to the \textit{Probabilistic AutoEncoder (PAE)} as shown in Fig. \ref{fig:ae}. The difference is that the dimension of the output layer of the network is twice the size of the input layer, which consists of $\bm{\mu}_{i}$ and $\bm{\sigma}_{i}^{2}$. A Softplus activation function is followed by the output variance to ensure positive values.
$\bm{\mu}_{i}$ represents the reconstruction result. $\bm{\sigma}_{i}^{2}$ can be termed as heteroscedastic aleatoric uncertainty, and has already established in the literature, as introduced in Section \ref{ssec:qu}. However, the intrinsic relationship between aleatoric uncertainty and AE-based OD has never been explained, and the potential of quantifying uncertainty in OD is not fully explored yet.}

	\textcolor{black}{PAE consists of two components, the encoder and decoder. The encoder transforms high-dimensional input data to low-dimensional latent variables, and the decoder does the opposite. Hence the process of PAE can be considered as a self-supervised two-steps regression. In this process, several input instances may be encoded into the same latent variable, then decoded into the same output. This phenomenon is caused by one of two reasons: either the randomness or noise inherent in inliers is eliminated during dimensionality reduction, or the pattern of outliers contradicts the knowledge learned by the model.
}

\textcolor{black}{After training, the reconstruction result fits the manifold of training data, while the estimated aleatoric uncertainty reflects the quality of the training set. Since the outputs and latent variables are in one-to-one correspondences, we can summarize the properties of the aleatoric uncertainty according to the value of the latent variables: when the data fall in the min-max range (determined by the training set) of the latent variables, the aleatoric uncertainty reflects multiple factors, including the randomness or noise of data, the region with absent data, and the region with sparse data (most likely outliers); when the data fall out of the min-max range of the latent variables, the aleatoric uncertainty becomes meaningless and elusive. More importantly, aleatoric uncertainty has a more positive impact when it is combined with reconstruction error. This suggests that we should focus on the form like NLL, leading to the weighted negative logarithmic likelihood:}

\textcolor{black}{
\begin{equation}
	\label{eq11}
	\text{WNLL}(\mathbf{x}, \bm{\mu}, \bm{\sigma}) = \sum_{d=1}^{D}{\alpha \frac{(x_{d}-\mu_{d})^{2}}{\sigma_{d}^{2}} + (1-\alpha) \ln \sigma_{d}^{2}},
\end{equation}
where $x_{d}$, $\mu_{d}$ and $\sigma_{d}$ denote the $d$-th attribute of a sample, $\mathbf{x}$, output mean values, $\bm{\mu}$, and output variance values, $\bm{\sigma}^{2}$, respectively. $\alpha \in [0,1]$ is a hyper-parameter that controls the trade-off between the two components of WNLL. Evidently, the first component is influenced by both reconstruction error and aleatoric uncertainty, while the second component is only influenced by aleatoric uncertainty. Moreover, aleatoric uncertainty has different effects on the two components. The smaller the value of $\alpha$, the stronger the positive effect of aleatoric uncertainty on WNLL. In this work, we use WNLL as the score function in the testing phase. By tuning $\alpha$, PAE adapts to the varying characteristics of the datasets. This delivers substantial benefits to various OD applications without extra modification to the training progress.}

\begin{figure}[t]
	\centering
	\centerline{\includegraphics[width=12.0cm]{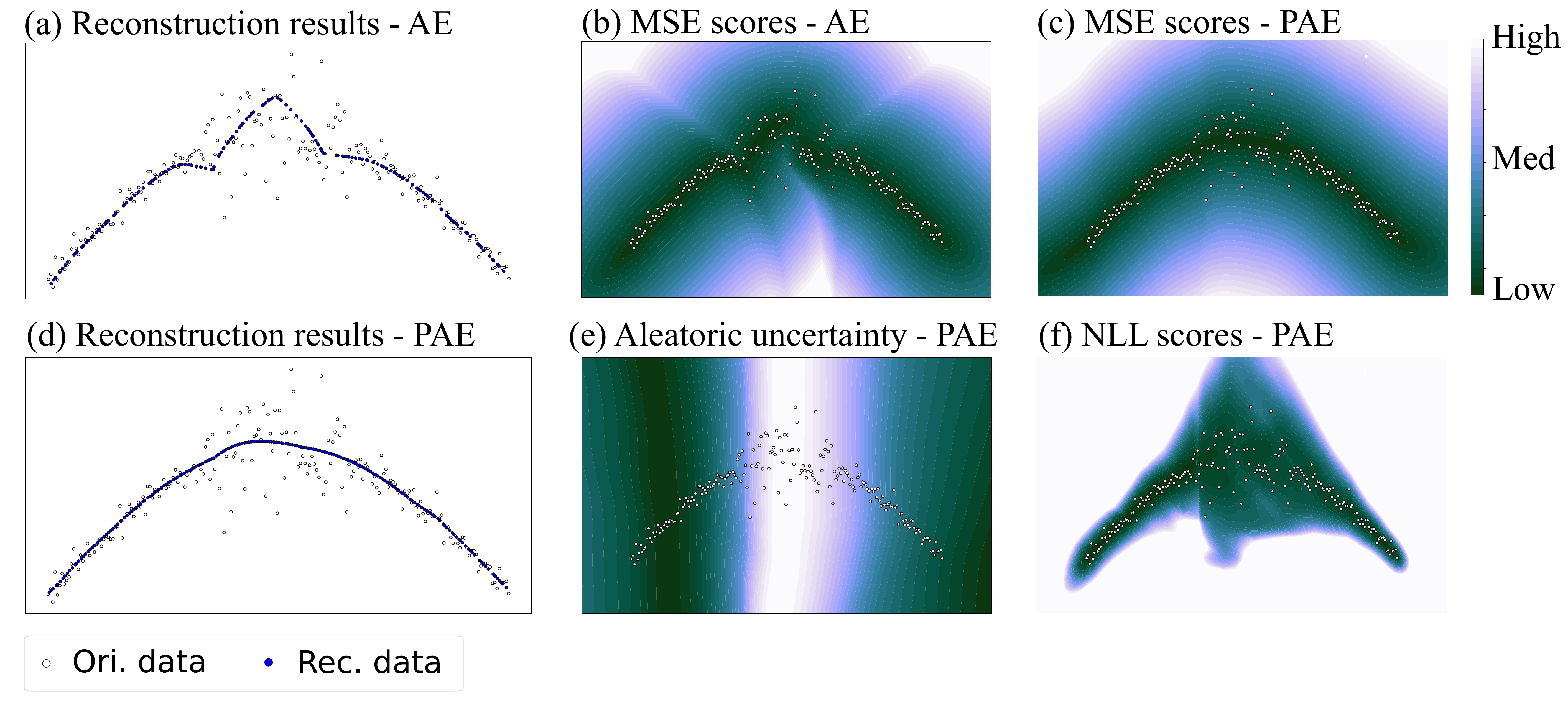}}
	\caption{\textcolor{black}{The first case contained a group of sinusoidal-shaped data with different Gaussian noise in different regions. (a) showed that the reconstruction results of AE were seriously affected by the noisy data. (b) showed that MSE scores computed from AE demonstrated the irregularity after reconstruction. (c) and (d) showed that the reconstruction results of PAE were smooth and essentially fit the original curve. (e) showed that PAE output high aleatoric uncertainty in the region between (-0.5,0.5) along horizontal axis. (f) showed that PAE assigned low NLL scores to the regions where data exist, but relatively high scores in the noisy regions.
			``Ori.'' and ``Rec.'' in the figures denote ``Original'' and ``Reconstruction'' respectively.}}
	\label{fig:example1}
\end{figure}

\subsection{Addressing the overconfidence with NLL and WNLL}
\label{ssec:ea}

\textcolor{black}{To verify the perspectives mentioned in the previous section and reveal how NLL and WNLL benefit OD, four cases were studied via visualization. AE and PAE were trained with same settings on each case. Since the cases are 2-dimensional and highly nonlinear, we set the number of units for each network layer for all models as [2,32,32,1,32,32,2], with scaled-exponential-linear-unit active function, except for the last layer.}

\textcolor{black}{The data in the first case came from a part of the sinusoidal curve, with different Gaussian noise in different regions. As shown in Fig. \ref{fig:example1}.(a), the original data are marked as white points, and the noise in the region between (-0.5,0.5) along horizontal axis is significantly larger than the other regions. We can observe that, the reconstruction results of AE (blue points), were seriously affected by the noisy data. MSE scores computed from AE also demonstrated the irregularity after reconstruction. However, this issue was substantially mitigated in PAE, which had a significantly improved reconstruction result that resembled the original curve, as shown in Fig. \ref{fig:example1}.(d). Additionally, Fig. \ref{fig:example1}.(e) showed that PAE output high aleatoric uncertainty in the region between (-0.5,0.5) along horizontal axis. This implied that the aleatoric uncertainty captured the noise in the original data. When reconstruction error and aleatoric uncertainty were considered simultaneously, we found that PAE assigned low NLL scores to the regions where data exist. This implied that PAE attempted to capture the distribution of the training set. On the other hand, the scores in the noisy regions were relatively high, caused by the high uncertainty. Thus this property is helpful for applications that treat noisy data as outliers.}

\begin{figure}[t]
	\centering
	\centerline{\includegraphics[width=12.0cm]{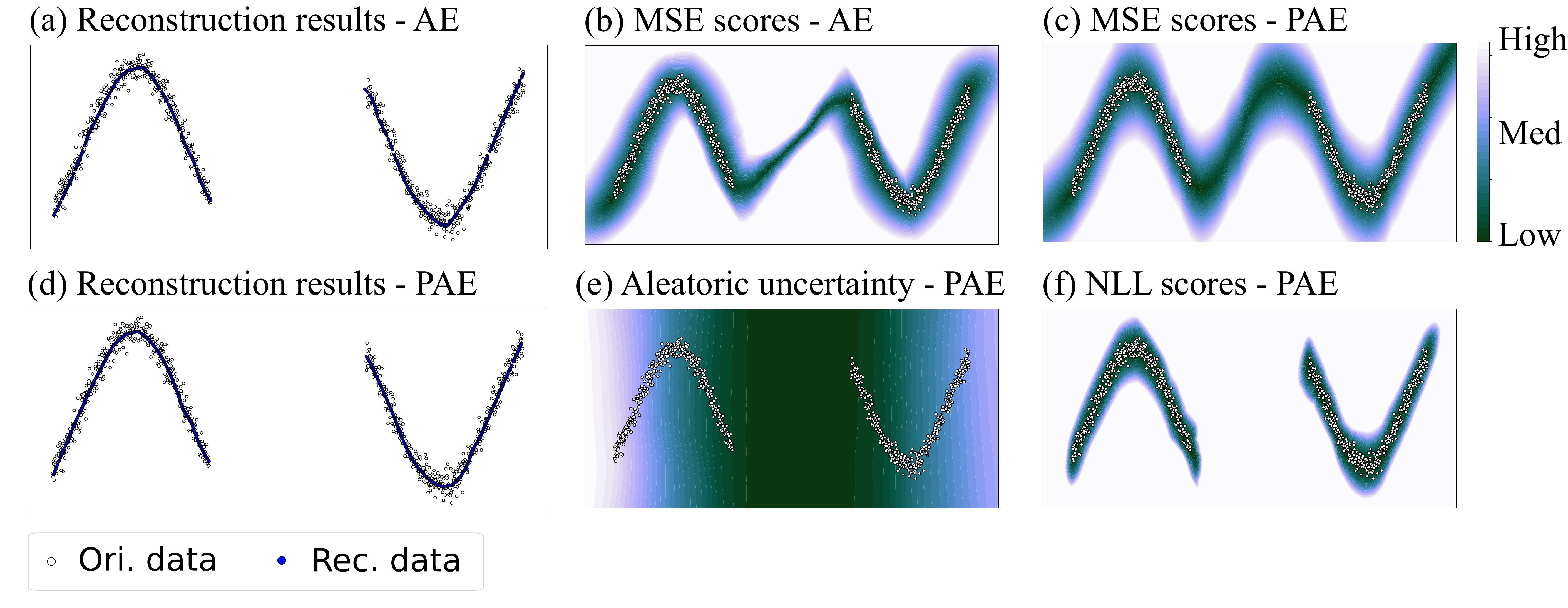}}
	\caption{\textcolor{black}{The second case contained two groups of sinusoidal-shaped data with equal density. (a) and (d) showed that AE and PAE both reconstructed the original data well. (b) and (c) showed that some regions between the two groups were also reconstructed well. (e) showed that PAE output low aleatoric uncertainty in the blank area between two groups. (f) showed that PAE only assigned low NLL scores to regions near the data.}}
	\label{fig:example2}
\end{figure}

\textcolor{black}{The second case contained two groups of sinusoidal-shaped data with equal density, as shown in Fig. \ref{fig:example2}. In this case, both AE and PAE reconstructed the original data well. However, some regions between the two groups (i.e., the hollow area) were also reconstructed with low error. This is unexpected since the models were required to learn only the pattern of data in the training set. This phenomenon is termed as model overconfidence, which is harmful to OD --- if we only focus on the reconstruction error. Fortunately, the aleatoric uncertainty estimated by PAE can solve the issue. Fig. \ref{fig:example2}.(e) showed that the aleatoric uncertainty became low in the blank area between two groups. Although some regions between the two groups had undesirable low reconstruction error, the final error could be enlarged by the low aleatoric uncertainty, according to the first term of Eq. \ref{eq10}. This result leaded to desirable NLL scores of PAE, as shown in Fig. \ref{fig:example2}.(f). In this way, the potential outliers in the blank area will not be misjudged.}

\begin{figure}[t]
\centering
\centerline{\includegraphics[width=12cm]{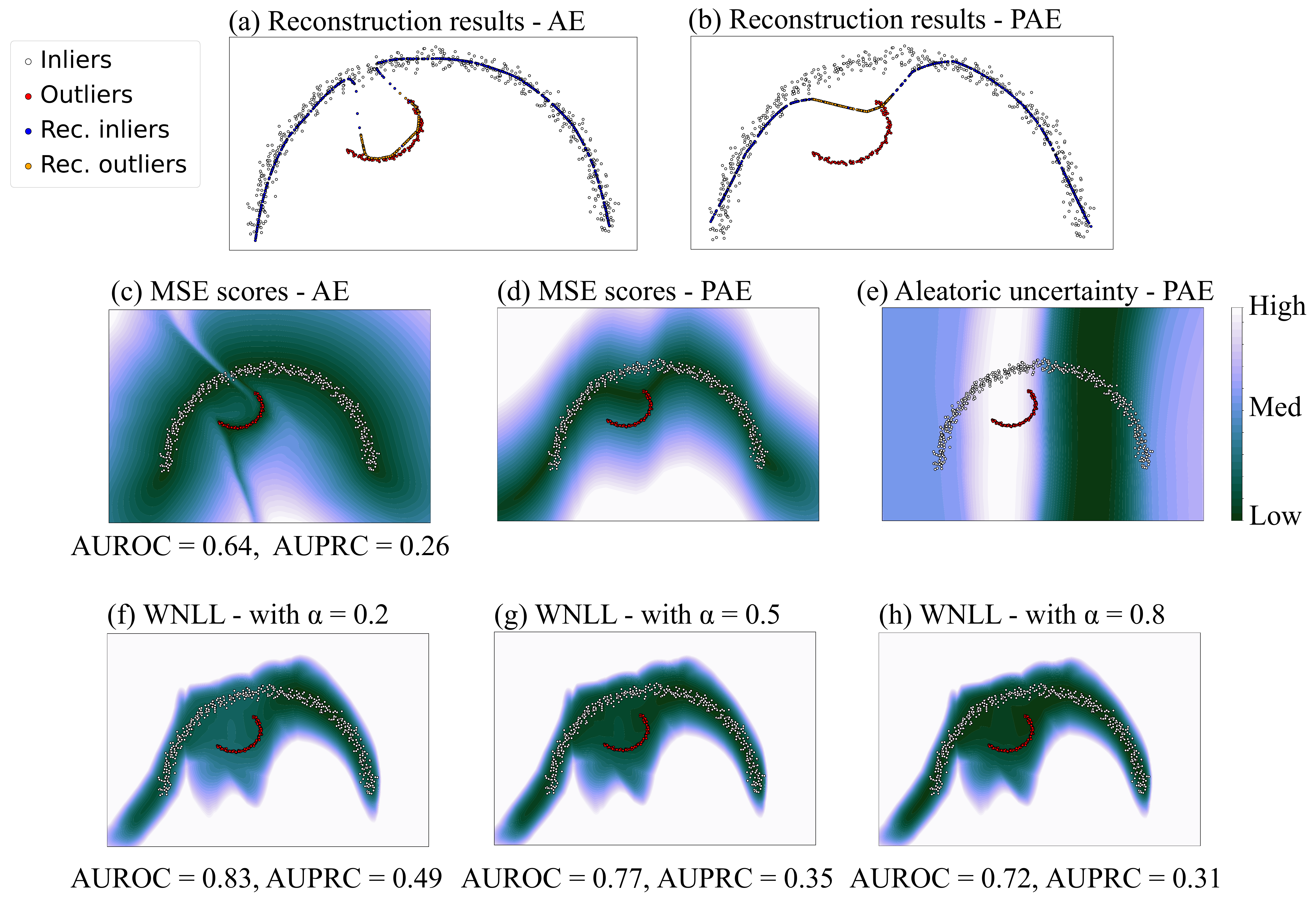}}
\caption{\textcolor{black}{The third cases included a large group of semicircular-shaped inliers and a small group of arcuate-shaped outliers, with an overlapping range on horizontal axis. (a) and (c) showed that AE mistakenly fit the outliers, resulting in bad OD performance. (b) and (d) showed that the reconstruction results of PAE did not deviate a lot from the inliers manifold. (e) showed that PAE outputs high aleatoric uncertainty around outliers, making outliers easier to be discriminated. (f) to (h) showed that PAE achieved better OD performance with smaller $\alpha$ in WNLL.}}
\label{fig:example3}
\end{figure}


\begin{figure}[t]
\centering
\centerline{\includegraphics[width=12cm]{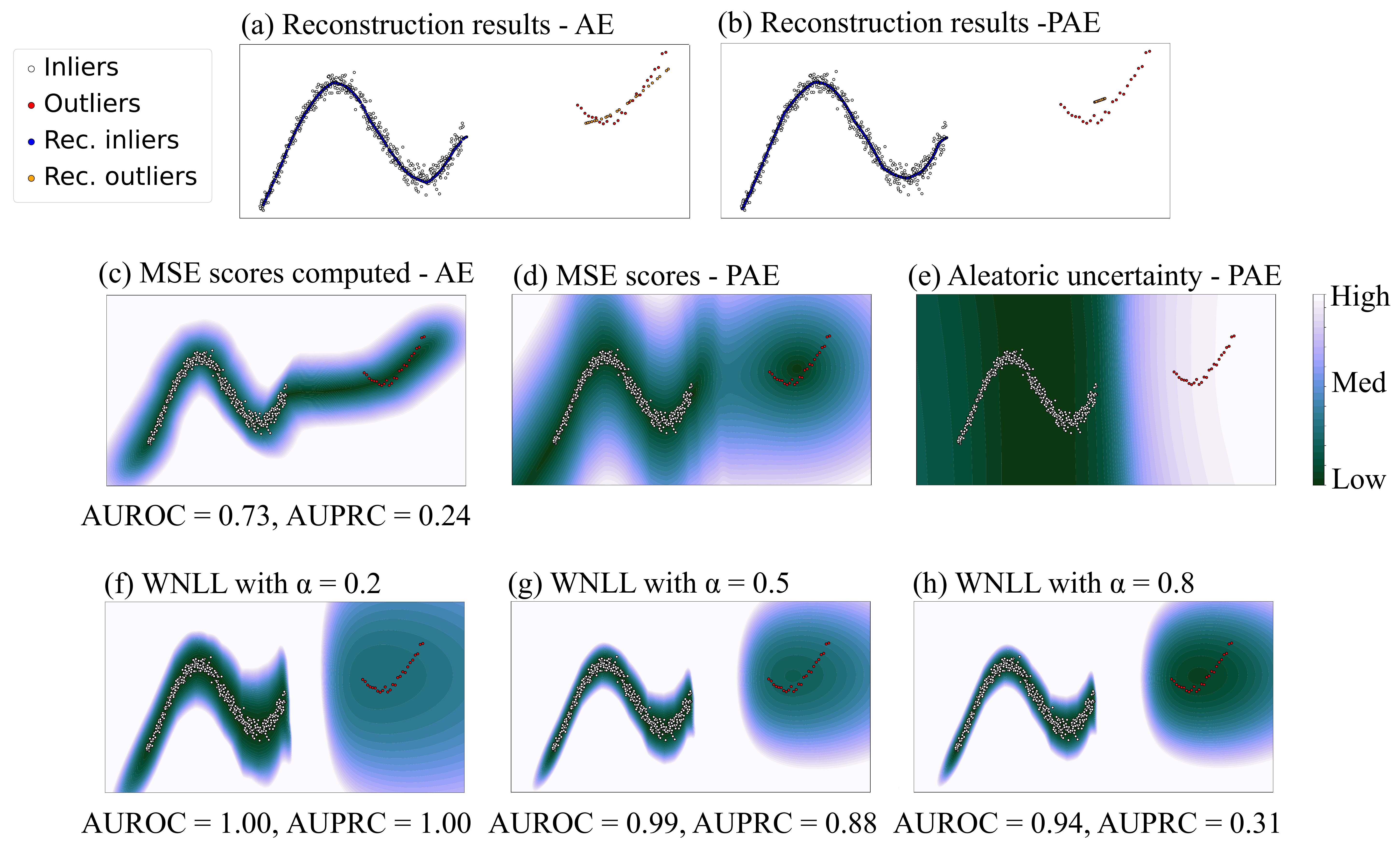}}
\caption{\textcolor{black}{The fourth cases includes a large group of sinusoidal-shaped inliers and a small group of arcuate-shaped outliers, with no overlapping on horizontal axis. (a) and (c) showed that AE reconstructed inliers well, but it output low MSE scores not only around outliers, but also between inliers and outliers. (b) and (d) showed that PAE did not reconstruct outliers well, but it still assigned relatively high MSE scores for some outliers. (e) showed that PAE output high aleatoric uncertainty in both of the aforementioned regions. (f) to (h) showed that PAE achieved better OD performance with smaller $\alpha$ in WNLL.}}
\label{fig:example4}
\end{figure}


\textcolor{black}{Data in the third and fourth case exhibited a large proportion of inliers and small proportion of collective outliers. Besides testing the reconstruction performance of AE and PAE, OD performance with different score functions was also tested. Area Under Receiver Operating characteristic Curve (AUROC) and Area Under Precision-Recall Curve (AUPRC) were used as the OD performance metrics. They were both suitable for label-unbalanced classification tasks, and ranged from 0 to 1, where 1 indicates the best performance.}

\textcolor{black}{The third case included a large group of semicircular-shaped inliers and a small group of arcuate-shaped outliers, with an overlapping range on the horizontal axis. As shown in Fig. \ref{fig:example3}, we can observe that AE was observed to erroneously fit outliers, resulting in poor OD performance. In contrast, PAE had different results. Its reconstruction results did not deviate significantly from the inliers. However, the overlap of inliers and outliers in parts of the latent space confused PAE, leading to regions with outliers having high aleatoric uncertainty. Therefore, even though PAE wanted to assign low NLL scores for all data, it could only assign relatively high scores for the regions with high concentration of outliers. In light of this, though some of the inliers were under suspicion, most outliers could be found out. To better prevent outliers from being erroneously identified as normal, we could decrease $\alpha$ in WNLL, since it improved the positive effect of aleatoric uncertainty. As a result, PAE achieved better OD performance when $\alpha$ = 0.2, compared with $\alpha$ = 0.5 (equivalent to NLL).}

\textcolor{black}{The last case included a large group of sinusoidal-shaped inliers and a small group of arcuate-shaped outliers, with no overlapping on the horizontal axis. As shown in Fig. \ref{fig:example4}, AE erroneously created regions with low MSE scores both around outliers, and between inliers and outliers, as shown in Fig. \ref{fig:example4}.(c). This phenomenon impaired OD performance. In contrast, with PAE, the sparsity and minority of outliers made PAE uncertain for the information in the region around outliers during training, thus output high aleatoric uncertainty there. For this situation, the positive effect of aleatoric uncertainty was essential for distinguishing, thus WNLL with smaller $\alpha$ yielded better OD performance.}

\textcolor{black}{Finally, we summarized the behaviors of aleatoric uncertainty for PAE as follows:}
\begin{enumerate}[1)]
\item \textcolor{black}{it increased in regions where data contained large randomness or noise;}
\item \textcolor{black}{it increased in regions where inliers and outliers overlapped in the latent space;}
\item \textcolor{black}{it increased in regions where data were sparse and few (most likely outliers);}
\item \textcolor{black}{it decreased in the hollow area within the data distribution.}
\end{enumerate}
\textcolor{black}{Moreover, we give some recommendations for the usage of NLL and WNLL in OD tasks:}
\begin{enumerate}[1)]
\item \textcolor{black}{NLL and WNLL were effective in mitigating the overconfidence issue when training data contained multiple inlier manifolds (patterns), and $\alpha$ in WNLL was recommended to set a high value here;}
\item \textcolor{black}{NLL and WNLL were effective when training data contained outliers, and $\alpha$ in WNLL was recommended to set a low value here.}
\item \textcolor{black}{For applications with both multiple inlier patterns and outlier contamination, practitioner should carefully adjust $\alpha$ in WNLL according to the specific analysis.}
\end{enumerate}

\subsection{AE is ignorance about the local relationship}

AE is trained to learn the normal patterns of the data, and outliers are intuitively expected to have higher reconstruction errors.
Although the reconstruction error such as MSE can be used as the outlier score directly, the score may be less reliable in certain situations. 
First, if the training dataset is contaminated by outliers, AE will be misled to learn abnormal patterns of the data, resulting in lower reconstruction errors of outliers and increasing difficulty of distinguishing outliers from inliers. 
Second, some outliers exhibit latent features conforming to the patterns of inliers, except that their feature values deviate from the normal range. These outliers may be unexpectedly well reconstructed by AE, making them difficult to distinguish from inliers.
\textcolor{black}{Although NLL and WNLL can alleviate the problem of outlier contamination, it is still necessary to search a complementary solution for any AE-based OD method.}

The issue raised can be explained by AEs' capturing global features of data, which lacked explicit consideration of local relationship. Therefore, to tackle this inadequacy, the local relationship of the data was considered as auxiliary information for AE. The mean-shift method could serve as a viable solution.
The concept of the mean-shift was first proposed by Fukunaga \textit{et al.} in 1975 \cite{fukunaga1975estimation}, applied to the estimation of the gradient of a density function. Then it was used in a variety of applications including clustering \cite{cheng1995mean}, image segmentation \cite{comaniciu2002mean}, and target tracking\cite{collins2003mean}. 

\begin{figure}[t]
\centering
\centerline{\includegraphics[width=9cm]{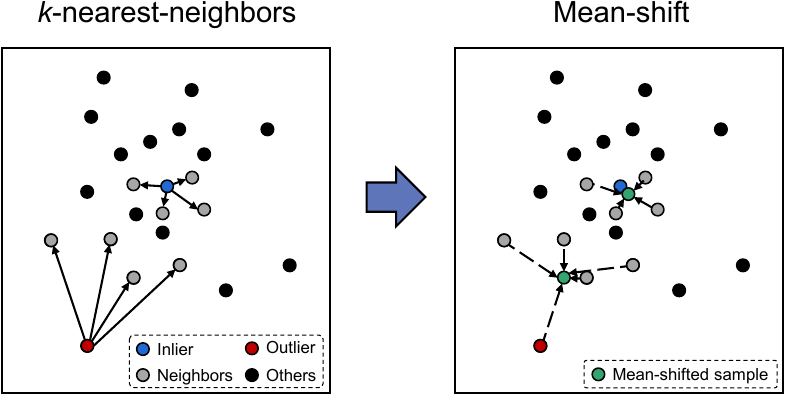}}

\caption{A 2-D illustration of the mean-shift with $m=1$ and $k=4$ for an inlier (top) and an outlier (bottom).
}
\label{fig:mod}
\end{figure}

The process of the mean-shift could be described as follows. Given a dataset $\mathbf{X} = \{\mathbf{x}_{1}, \mathbf{x}_{2}, \cdots, \mathbf{x}_{N}\}$, mean-shift first calculated each sample's $k$-nearest-neighbors in $\mathbf{X}$. For example, for the sample $\mathbf{x}_{i}$, its $k$-nearest-neighbors set was $\text{NL}_{i}=\{\mathbf{x}_{i}^{1}, \mathbf{x}_{i}^{2}, \cdots, \mathbf{x}_{i}^{k}\}$. Mean-shift then added the sample itself to the neighboring set $\text{NL}_{i}=\{\mathbf{x}_{i}, \mathbf{x}_{i}^{1}, \mathbf{x}_{i}^{2}, \cdots, \mathbf{x}_{i}^{k}\}$ and calculated the mean value of the list, as expressed in Eq. \ref{eq3}.
\begin{equation}
\label{eq3}
\left\{
\begin{aligned}
	& \mathbf{x}_{i}^{{\rm MS}(1,k)} = \frac{1}{|\text{NL}_{i}|} \sum_{\mathbf{x} \in \text{NL}_{i}} {\mathbf{x}}, \\
	& \mathbf{X}^{{\rm MS}(1,k)} = \{\mathbf{x}_{1}^{{\rm MS}(1,k)}, \mathbf{x}_{2}^{{\rm MS}(1,k)}, \cdots, \mathbf{x}_{N}^{{\rm MS}(1,k)}\}.
\end{aligned}
\right.
\end{equation}
This process can be repeated for several times to get a more compact data distribution.
Specifically, the same process could be applied to $\mathbf{X}^{{\rm MS}(1,k)}$ to get $\mathbf{X}^{{\rm MS}(2,k)}$, in which ${\rm MS}(m,k)$ denotes the process of mean-shift with $k$ neighbors and $m$ shift times. A 2-D illustration of mean-shift is shown in Fig. \ref{fig:mod}.

To apply the mean-shift on the AE-based OD method, the following procedure was proposed: the AE could be trained using the loss function  Eq. \ref{eq9} or Eq. \ref{eq10} during the training phase. During the scoring phase, the outlier score of a test sample $\mathbf{x}$ could be computed using the score function as illustrated in Eq. \ref{eq12} and Eq. \ref{eq13}. 
\begin{align}
\label{eq12}
&\text{MSS-MSE}(\mathbf{x}) = ||\mathbf{x}^{{\rm MS}(m,k)} - \hat{\mathbf{x}}||_{2}^{2}, \\
&\text{or}, \nonumber \\
\label{eq13}
&\text{MSS-WNLL}(\mathbf{x}) = \sum_{d=1}^{D}{\alpha \frac{(x_{d}^{{\rm MS}(m,k)}-\mu_{d})^{2}}{\sigma_{d}^{2}} + (1-\alpha) \ln \sigma_{d}^{2}},
\end{align}
where $\mathbf{x}^{{\rm MS}(m,k)}$ denotes the mean-shifted result of the test sample $\mathbf{x}$.

\subsection{Excluding well-reconstructed outliers using MSS}

The concept of MSS can be analyzed from two perspectives. 
Upon analysis of the distance measurement, the reconstruction error could be viewed as the deviation of the object's position in the original data space. The reconstruction error of the inlier was relatively small if AE was well-trained, which means the reconstructed inlier lies within close proximity to the original position. Meanwhile, the mean-shifted result of an inlier was also close to its original position, hence the distance between the reconstructed inlier and the mean-shifted inlier would be small. For the outliers, the mean-shifted result was generally closer to inliers and further from its original locality. The outliers' reconstruction result deviated in a larger magnitude for distance, and with a random direction, since AE learning was inadequate. Thus, the distance between an outlier's mean-shifted result and reconstruction result would be significantly large compared to an inlier, which made them simple to distinguish.
Upon analysis of the reconstruction likelihood, even though an outlier was reconstructed with a relatively high likelihood, its mean-shifted result was not similar to the reconstruction. This resulted in a high outlier score as the mean-shifted result fell into a position with low likelihood within the likelihood distribution.

In theory, usage of the distance between the mean of $k$-NN instead of the original data and the reconstructed data had two benefits. First, some data shared common mean values, hence similar data in the feature space could have similar outlier scores \cite{yang2020outlier}. Therefore, it avoided local variance in outlier score space. Second, the mean values were seen as the representatives of $k$-NN, thus it considered both object-level and group-level factors when scoring \cite{yang2022neighborhood}. Therefore, it provided a robust method for the detection of both point outliers and collective outliers.

\begin{figure}[t]
\centering
\centerline{\includegraphics[width=12.0cm]{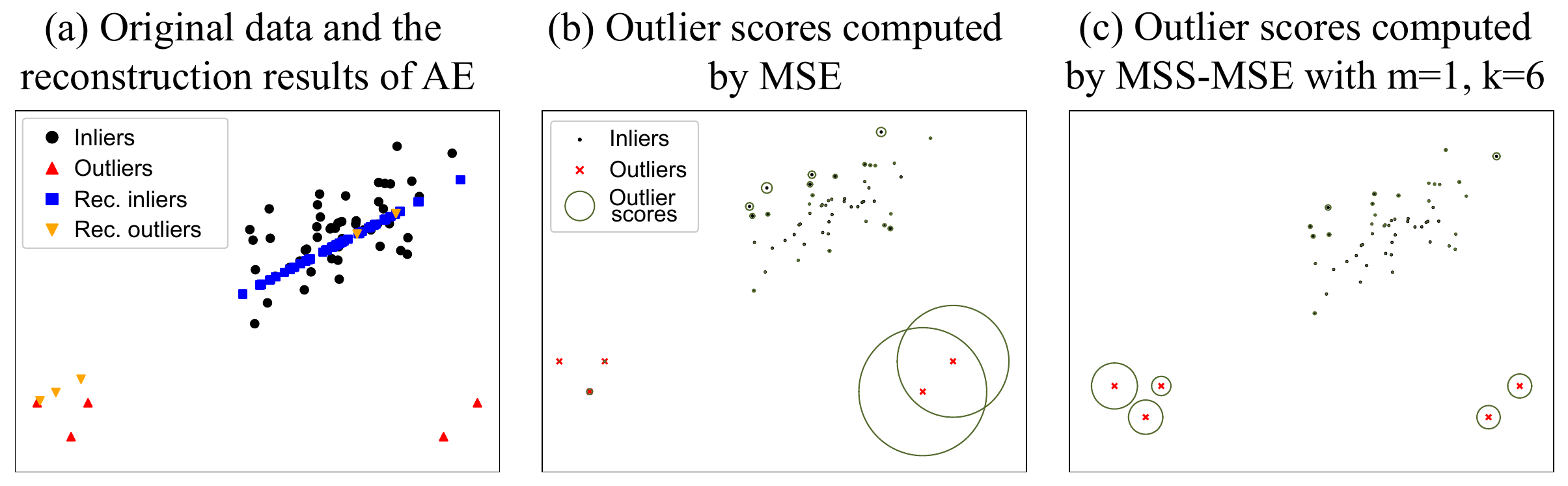}}
\caption{A 2-D example of the effect of the MSS. An AE was trained using these data with the loss function of MSE. (b) and (c) showed the outlier scores generated by MSE and MSS-MSE with $m=1, k=6$ respectively. The larger the radius of the green circle, the higher the outlier score.}
\label{fig:ae2d}
\end{figure}


\begin{figure}[t]
\centerline{\includegraphics[width=12cm]{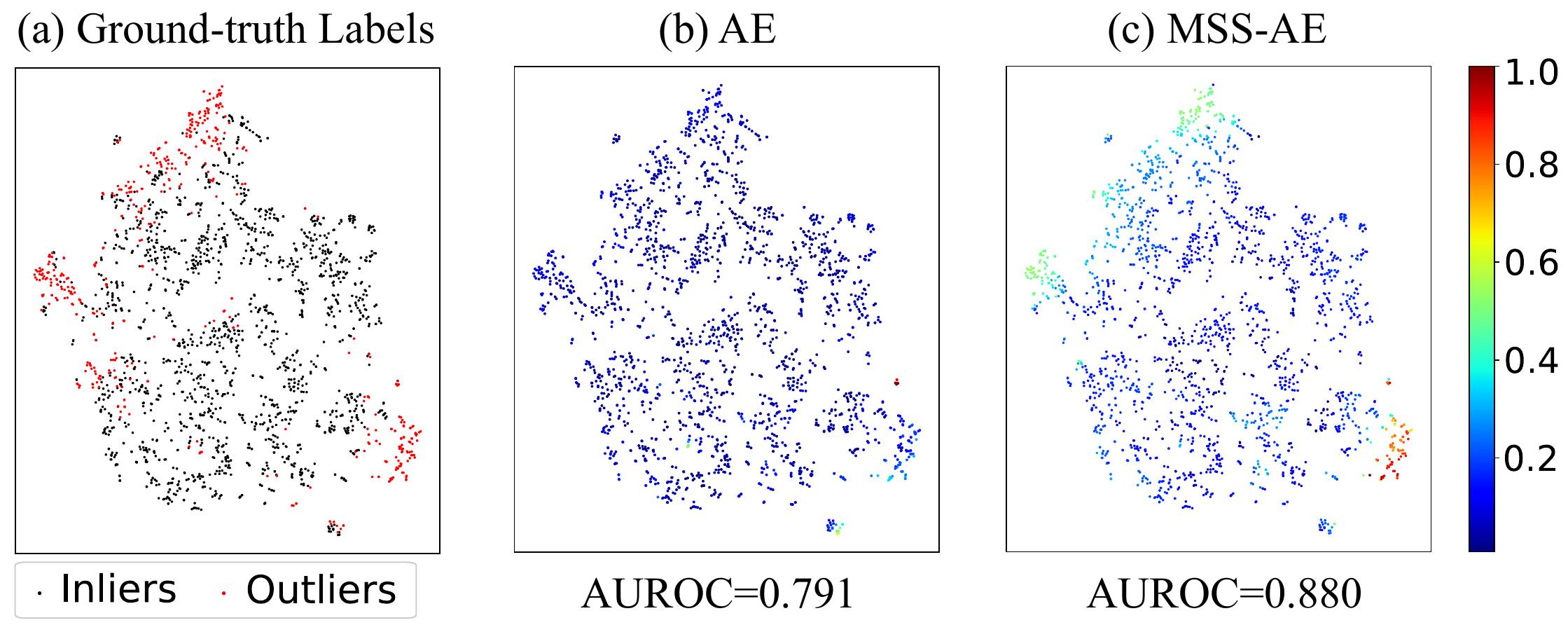}}
\caption{An illustration of the MSS method on the \textit{Cardiotocography} dataset. The data in \textit{Cardiotocography} were dimension-reduced to 2-D utilizing t-SNE. (a) showed the ground-truth label map of the data, in which the black points denote the inliers and the red points denote the outliers. (b) and (c) showed the outlier score map generated by AE and MSS-AE respectively.}

\label{fig:vis}
\end{figure}

An example of the effect of MSS was shown in Fig. \ref{fig:ae2d}. The outliers lying on the bottom left were exactly situated along the feature direction of inliers, thus their reconstruction errors were unexpectedly small, making identification of these outliers from inliers challenging. However, since they were far way from inliers, they could easily be identified after applying MSS with $m = 1$ and $k = 6$.

\begin{figure}[t]
\centering
\centerline{\includegraphics[width=10cm]{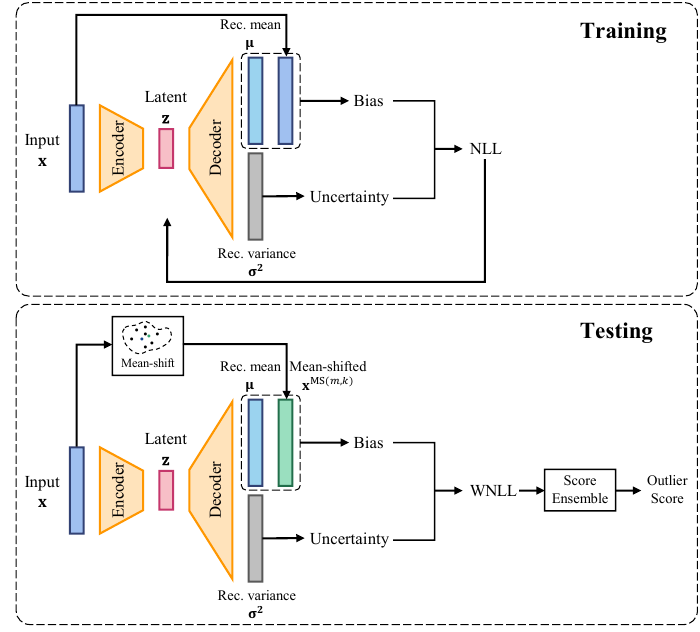}}

\caption{An illustration of the training and testing procedure of MSS-PAE. ``Rec.'' in the figure denotes ``Reconstruction''.
}
\label{fig:proc}
\end{figure}

Fig. \ref{fig:vis} demonstrated an example of the MSS method on the \textit{Cardiotocography} dataset. In this case, MSS was applied to the conventional AE, termed MSS-AE. For better presentation, the data in \textit{Cardiotocography} were dimension-reduced to 2-D utilizing t-distributed stochastic neighbor embedding (t-SNE) \cite{van2008visualizing}. Fig. \ref{fig:vis}. (a) showed the ground-truth label map of the data, where the black points denoted the inliers and the red points denoted the outliers. The outliers were observed to be distributed in the top left region and bottom right region of the map. Figs. \ref{fig:vis}. (b) and (c) illustrated the outlier score map generated by AE and MSS-AE respectively. The color scale corresponded to the outlier score of each object. The performance was evaluated by AUROC. 
AE could distinguish the part of the outliers lying at the bottom right and achieve a AUROC of 0.791, but could not distinguish the outliers at the top left either. The outlier scores were also not differentiable enough. In contrast, MSS-AE could distinguish both outliers lying at the bottom right and top left. MSS-AE provided more differentiable outlier scores and a significant improvement in AUROC performance to 0.880.

MSS could also be applied to PAE, producing MSS-PAE. An illustration of the training and testing procedure of MSS-PAE was shown in Fig. \ref{fig:proc}. The performance of MSS-PAE was evaluated in the Section \ref{sec:exp}.

\section{Experiments}
\label{sec:exp}
In this work, empirical experiments were conducted on 32 commonly used real-world outlier detection datasets. The proposed methods were compared with 5 typical AE-based OD methods (including AE itself), and proved the efficacy of applying MSS. A comparison was made against 8 classic non-AE-based SOTA OD methods.

The experiments answered the following questions:

\noindent1) How effective and flexible is WNLL in real-world scenarios? (Section \ref{sssec:ep})

\noindent2) How do hyper-parameters $k$ and $m$ affect MSS? (Section \ref{sssec:ems})

\noindent3) Is MSS beneficial for other AE-based methods? (Section \ref{sssec:ca})

\noindent4) Comparing the proposed methods with 5 AE-based and 8 non-AE-based OD methods, which one yields the best performance? (Section \ref{sssec:ca})

\subsection{Datasets and baselines}
\label{ssec:dat}
32 real-world outlier detection datasets were obtained from DAMI\footnote{DAMI: www.dbs.ifi.lmu.de/research/outlier-evaluation/DAMI} and ODDS\footnote{ODDS: odds.cs.stonybrook.edu} dataset repositories. The summary of these datasets was shown in TABLE \ref{tab:ds}. We divided each dataset into the training set and test set with a ratio of 3:1. Then, one-third of the samples in the training set were selected to form the validation set. The outlier ratios were kept constant in all three sets, and the validation set and test set included the data labels but the training set did not. In addition, each attribute of the data was normalized to zero-mean and unit-variance using the Mean-SD normalization \cite{kandanaarachchi2020normalization}. The mean value and the variance value were obtained from the training set. This operation was essential for the AE-based methods because it balanced the scale of each attribute of the data, lest the attributes with relatively large scale of values had unexpectedly bigger impacts on the model optimization.

To show the superiority of the proposed methods, 5 typical AE-based methods and 8 non-AE-based SOTA OD methods were used as the baselines. The AE-based methods included AE \cite{hawkins2002outlier}, RDA \cite{zhou2017anomaly}, LCAE \cite{ishii2019low}, RSRAE \cite{yu2021autoencoder} and FDAE \cite{guo2022unsupervised}. The details of these methods were introduced in Section \ref{ssec:tae}. The non-AE-based OD methods included proximity-based Local Outlier Factor (LOF) \cite{breunig2000lof}, Mean-shift Outlier Detector (MOD) \cite{yang2021mean}, and Density-increasing Path (DIP) \cite{zhao2023searching}, classification-based One-Class Support Vector Machines (OCSVM) \cite{scholkopf2001estimating}, ensemble-based Isolation Forest (iForest) \cite{liu2008isolation} and Deep Isolation Forest (DIF) \cite{xu2023deep}, probabilistic-based Histogram-based Outlier Score (HBOS) \cite{goldstein2012histogram} and empirical-Cumulative-distribution-based Outlier Detection (ECOD) \cite{li2022ecod}. The AE-based methods were reproduced, compared to their original results, and the proposed MSS method was applied to improve the performance. All the methods were implemented using Python, and the implementation from PyOD \cite{zhao2019pyod} was utilized for non-AE-based methods except DIP, which was implemented using its original codes. AUROC was used as the evaluation metric. 

\begin{table}[t]
	\centering
	\caption{Information of 32 real-world outlier detection datasets.}
	\label{tab:ds}
	\scalebox{0.8}{
		\begin{tabular}{c|cccc}
			\toprule
			\textbf{Name}    & \textbf{Dimension} & \textbf{Instances} & \textbf{Outliers} & \textbf{Ratio} \\ \midrule
			ALOI             & 27           & 49,534             & 1,508             & 3.0\%          \\
			Annthyroid       & 21           & 7,129              & 534               & 7.5\%          \\
			Arrhythmia       & 259          & 450                & 206               & 45.8\%         \\
			Breastw          & 9            & 683                & 239               & 35.0\%         \\
			Cardiotocography & 21           & 2,114              & 466               & 22.0\%         \\
			Glass            & 7            & 214                & 9                 & 4.2\%          \\
			HeartDisease     & 13           & 270                & 120               & 44.4\%         \\
			InternetAds      & 1,555        & 1,966              & 368               & 18.7\%         \\
			Ionosphere       & 32           & 351                & 126               & 35.9\%         \\
			Letter           & 32           & 1,600              & 100               & 6.3\%          \\
			Lymphography     & 3            & 148                & 6                 & 4.1\%          \\
			Mammography      & 6            & 11,183             & 260               & 2.3\%          \\
			Mnist            & 100          & 7,603              & 700               & 9.2\%          \\
			Musk             & 166          & 3,062              & 97                & 3.2\%          \\
			Optdigits        & 64           & 5,216              & 150               & 2.9\%          \\
			PageBlocks       & 10           & 5,393              & 510               & 9.5\%          \\
			Parkinson        & 22           & 195                & 147               & 75.4\%         \\
			PenDigits        & 16           & 9,868              & 20                & 0.2\%          \\
			Pima             & 8            & 768                & 268               & 34.9\%         \\
			Satellite        & 36           & 6,435              & 2,036             & 31.6\%         \\
			Satimage-2       & 36           & 5,803              & 71                & 1.2\%          \\
			Shuttle          & 9            & 1,013              & 13                & 1.3\%          \\
			SpamBase         & 57           & 4,207              & 1,679             & 39.9\%         \\
			Speech           & 400          & 3,686              & 61                & 1.7\%          \\
			Stamps           & 9            & 340                & 31                & 9.1\%          \\
			Thyroid          & 6            & 3,772              & 93                & 2.5\%          \\
			Vertebral        & 6            & 240                & 30                & 12.5\%         \\
			Vowels           & 12           & 1,456              & 50                & 3.4\%          \\
			Waveform         & 21           & 3,443              & 100               & 2.9\%          \\
			Wilt             & 5            & 4,819              & 257               & 5.3\%          \\
			Wine             & 13           & 129                & 10                & 7.8\%          \\
			WPBC             & 33           & 198                & 47                & 23.7\%         \\ \bottomrule
		\end{tabular}
	}
\end{table}

\subsection{Experimental setup}
\label{ssec:set}

\subsubsection{\textcolor{black}{Network architecture}}
The architecture of PAE had been introduced in Section \ref{ssec:apre}. The model used NLL as the training loss function and WNLL as the scoring function.
MSS is a universal scoring method that can be applied to most AE-based methods including the proposed PAE. As notations, application to AE produced MSS-AE and application to RSRAE produced MSS-RSRAE, etc. 

\begin{table}[t]
	\centering
	\caption{Architecture setting of the AE-based methods.}
	\label{tab:ns}
	\scalebox{0.8}{
		\begin{tabular}{c|c|c}
			\toprule
			\textbf{Dimsension (D)}            & \textbf{Number of layers} & \textbf{Number of units}                                                                                        \\ \midrule
			\textless{}20             & 3                & {[}D, D // 2, D{]}                                                                                     \\ \midrule
			$\geq$20, \textless{}100  & 5                & \begin{tabular}[c]{@{}c@{}}{[}D, D // 2,\\ D // 4,\\ D // 2, D{]}\end{tabular}                     \\ \midrule
			$\geq$100, \textless{}200 & 7                & \begin{tabular}[c]{@{}c@{}}{[}D, D // 2, D // 4,\\ D // 8,\\ D // 4, D // 2, D{]}\end{tabular} \\ \midrule
			$\geq$200                 & 7                & \begin{tabular}[c]{@{}c@{}}{[}D, D // 2, D // 4,\\ D // 16,\\ D // 4, D // 2, D{]}\end{tabular} \\ \bottomrule
	\end{tabular}}
\end{table}

Considering all data were tabular data, fully-connected layers were used to construct the AE. The number of the layers depended on the dimension of the input data. The detailed setting of the network architecture was shown in TABLE \ref{tab:ns}. Then, the rectified linear unit (ReLU) activation function was added after each layer of the AE except the last layer to increase the nonlinearity of the network. ReLU is widely used as an activation function in deep neural networks \cite{ramachandran2017searching}. It mimics neuronal mechanisms in the human brain, and demonstrated good performance in many tasks while requiring minimal computational resources. The settings were applied to all the AE-based methods in this experiment. 

\begin{table}[t]
\centering
\caption{AUROC results of AE and PAE with different $\alpha$.}
\label{tab:pae}
\scalebox{0.75}{
	\begin{tabular}{c|c|ccccc}
		\toprule
		\multirow{2}{*}{\textbf{Dataset}} & \multirow{2}{*}{\textbf{AE}} & \multicolumn{5}{c}{\textbf{PAE}* with different $\alpha$}                                                                                                                                                                   \\ \cline{3-7} 
		&                              & \multicolumn{1}{c|}{0.20} & \multicolumn{1}{c|}{0.33} & \multicolumn{1}{c|}{0.50} & \multicolumn{1}{c|}{0.66} & 0.80 \\ \midrule
		ALOI                              & 0.563                       & \multicolumn{1}{c|}{0.545}                & \multicolumn{1}{c|}{0.559}                & \multicolumn{1}{c|}{0.562}              & \multicolumn{1}{c|}{0.556}                & \textbf{0.564}       \\
		Annthyroid                        & 0.608                        & \multicolumn{1}{c|}{0.799}                & \multicolumn{1}{c|}{0.793}                & \multicolumn{1}{c|}{0.790}              & \multicolumn{1}{c|}{\textbf{0.850}}       & 0.835                \\
		Arrhythmia                        & 0.714                        & \multicolumn{1}{c|}{0.736}                & \multicolumn{1}{c|}{\textbf{0.740}}       & \multicolumn{1}{c|}{0.624}              & \multicolumn{1}{c|}{0.706}                & 0.649                \\
		Breastw                           & 0.972                        & \multicolumn{1}{c|}{0.994}                & \multicolumn{1}{c|}{\textbf{0.995}}       & \multicolumn{1}{c|}{0.994}              & \multicolumn{1}{c|}{0.971}                & 0.971                \\
		Cardiotocography                  & 0.811                        & \multicolumn{1}{c|}{0.818}                & \multicolumn{1}{c|}{0.818}                & \multicolumn{1}{c|}{0.819}              & \multicolumn{1}{c|}{0.858}                & \textbf{0.870}       \\
		Glass                             & 0.804                        & \multicolumn{1}{c|}{0.961}                & \multicolumn{1}{c|}{\textbf{0.980}}       & \multicolumn{1}{c|}{0.873}              & \multicolumn{1}{c|}{0.912}                & 0.882                \\
		HeartDisease                      & 0.813                        & \multicolumn{1}{c|}{0.865}                & \multicolumn{1}{c|}{\textbf{0.871}}       & \multicolumn{1}{c|}{0.859}              & \multicolumn{1}{c|}{0.831}                & 0.768                \\
		InternetAds                       & \textbf{0.706}               & \multicolumn{1}{c|}{0.685}                & \multicolumn{1}{c|}{0.681}                & \multicolumn{1}{c|}{0.671}              & \multicolumn{1}{c|}{0.674}                & 0.697                \\
		Ionosphere                        & 0.971                        & \multicolumn{1}{c|}{\textbf{0.985}}       & \multicolumn{1}{c|}{0.971}                & \multicolumn{1}{c|}{0.914}              & \multicolumn{1}{c|}{0.957}                & 0.956                \\
		Letter                            & \textbf{0.857}               & \multicolumn{1}{c|}{0.855}                & \multicolumn{1}{c|}{0.768}                & \multicolumn{1}{c|}{0.808}              & \multicolumn{1}{c|}{0.764}                & 0.807                \\
		Lymphography                      & \textbf{0.971}               & \multicolumn{1}{c|}{0.914}                & \multicolumn{1}{c|}{0.914}                & \multicolumn{1}{c|}{0.914}              & \multicolumn{1}{c|}{0.914}                & 0.914                \\
		Mammography                       & \textbf{0.893}               & \multicolumn{1}{c|}{0.884}                & \multicolumn{1}{c|}{0.882}                & \multicolumn{1}{c|}{0.891}              & \multicolumn{1}{c|}{0.877}                & 0.874                \\
		Mnist                             & 0.882                        & \multicolumn{1}{c|}{\textbf{0.948}}       & \multicolumn{1}{c|}{0.945}                & \multicolumn{1}{c|}{0.931}              & \multicolumn{1}{c|}{0.906}                & 0.891                \\
		Musk                              & \textbf{1.000}               & \multicolumn{1}{c|}{\textbf{1.000}}       & \multicolumn{1}{c|}{\textbf{1.000}}       & \multicolumn{1}{c|}{\textbf{1.000}}     & \multicolumn{1}{c|}{\textbf{1.000}}       & \textbf{1.000}       \\
		Optdigits                         & 0.760                        & \multicolumn{1}{c|}{\textbf{0.936}}       & \multicolumn{1}{c|}{0.912}                & \multicolumn{1}{c|}{0.865}              & \multicolumn{1}{c|}{0.850}                & 0.874                \\
		PageBlocks                        & 0.952                        & \multicolumn{1}{c|}{\textbf{0.962}}       & \multicolumn{1}{c|}{0.946}                & \multicolumn{1}{c|}{0.949}              & \multicolumn{1}{c|}{0.942}                & 0.953                \\
		Parkinson                         & 0.627                        & \multicolumn{1}{c|}{\textbf{0.815}}       & \multicolumn{1}{c|}{0.662}                & \multicolumn{1}{c|}{0.711}              & \multicolumn{1}{c|}{0.611}                & 0.576                \\
		PenDigits                         & 0.879                        & \multicolumn{1}{c|}{\textbf{0.951}}       & \multicolumn{1}{c|}{0.929}                & \multicolumn{1}{c|}{0.913}              & \multicolumn{1}{c|}{0.897}                & 0.907                \\
		Pima                              & 0.651                        & \multicolumn{1}{c|}{\textbf{0.756}}       & \multicolumn{1}{c|}{0.736}                & \multicolumn{1}{c|}{0.698}              & \multicolumn{1}{c|}{0.696}                & 0.683                \\
		Satellite                         & 0.750                        & \multicolumn{1}{c|}{\textbf{0.855}}       & \multicolumn{1}{c|}{0.845}                & \multicolumn{1}{c|}{0.817}              & \multicolumn{1}{c|}{0.777}                & 0.740                \\
		Satimage-2                        & \textbf{0.993}               & \multicolumn{1}{c|}{0.983}                & \multicolumn{1}{c|}{0.983}                & \multicolumn{1}{c|}{0.988}              & \multicolumn{1}{c|}{0.988}                & 0.988                \\
		Shuttle                           & 0.995                        & \multicolumn{1}{c|}{\textbf{0.996}}       & \multicolumn{1}{c|}{0.988}                & \multicolumn{1}{c|}{0.992}              & \multicolumn{1}{c|}{0.992}                & 0.992                \\
		SpamBase                          & 0.565                        & \multicolumn{1}{c|}{\textbf{0.847}}       & \multicolumn{1}{c|}{0.793}                & \multicolumn{1}{c|}{0.712}              & \multicolumn{1}{c|}{0.646}                & 0.650                \\
		Speech                            & 0.551                        & \multicolumn{1}{c|}{0.541}                & \multicolumn{1}{c|}{0.532}                & \multicolumn{1}{c|}{0.561}              & \multicolumn{1}{c|}{\textbf{0.597}}       & 0.495                \\
		Stamps                            & 0.931                        & \multicolumn{1}{c|}{0.844}                & \multicolumn{1}{c|}{\textbf{0.952}}       & \multicolumn{1}{c|}{0.950}              & \multicolumn{1}{c|}{0.948}                & 0.948                \\
		Thyroid                           & 0.975                        & \multicolumn{1}{c|}{0.973}                & \multicolumn{1}{c|}{\textbf{0.991}}       & \multicolumn{1}{c|}{0.977}              & \multicolumn{1}{c|}{0.979}                & 0.982                \\
		Vertebral                         & 0.607                        & \multicolumn{1}{c|}{0.676}                & \multicolumn{1}{c|}{\textbf{0.772}}       & \multicolumn{1}{c|}{0.349}              & \multicolumn{1}{c|}{0.747}                & 0.758                \\
		Vowels                            & 0.937                        & \multicolumn{1}{c|}{0.968}                & \multicolumn{1}{c|}{\textbf{0.976}}       & \multicolumn{1}{c|}{0.972}              & \multicolumn{1}{c|}{0.955}                & 0.931                \\
		Waveform                          & 0.781                        & \multicolumn{1}{c|}{\textbf{0.881}}       & \multicolumn{1}{c|}{0.790}                & \multicolumn{1}{c|}{0.776}              & \multicolumn{1}{c|}{0.771}                & 0.771                \\
		Wilt                              & 0.383                        & \multicolumn{1}{c|}{0.822}                & \multicolumn{1}{c|}{0.823}                & \multicolumn{1}{c|}{0.818}              & \multicolumn{1}{c|}{0.861}                & \textbf{0.879}       \\
		Wine                              & \textbf{0.914}               & \multicolumn{1}{c|}{0.879}                & \multicolumn{1}{c|}{0.897}                & \multicolumn{1}{c|}{0.897}              & \multicolumn{1}{c|}{\textbf{0.914}}       & \textbf{0.914}       \\
		WPBC                              & 0.499                        & \multicolumn{1}{c|}{\textbf{0.548}}       & \multicolumn{1}{c|}{0.516}                & \multicolumn{1}{c|}{0.501}              & \multicolumn{1}{c|}{0.484}                & 0.479                \\ \midrule
		AVG.                     & 0.791                        & \multicolumn{1}{c|}{\textbf{0.851}}       & \multicolumn{1}{c|}{0.843}                & \multicolumn{1}{c|}{0.815}              & \multicolumn{1}{c|}{0.826}                & 0.819                \\ 
		BEST & 7                        & \multicolumn{1}{c|}{\textbf{13}}       & \multicolumn{1}{c|}{9}                & \multicolumn{1}{c|}{1}              & \multicolumn{1}{c|}{3}                & 5
		\\ \bottomrule
	\end{tabular}
}
\begin{tablenotes}
	\footnotesize
	\item * denotes the proposed method.
\end{tablenotes}
\end{table}

\begin{figure}[t]
\centerline{\includegraphics[width=12cm]{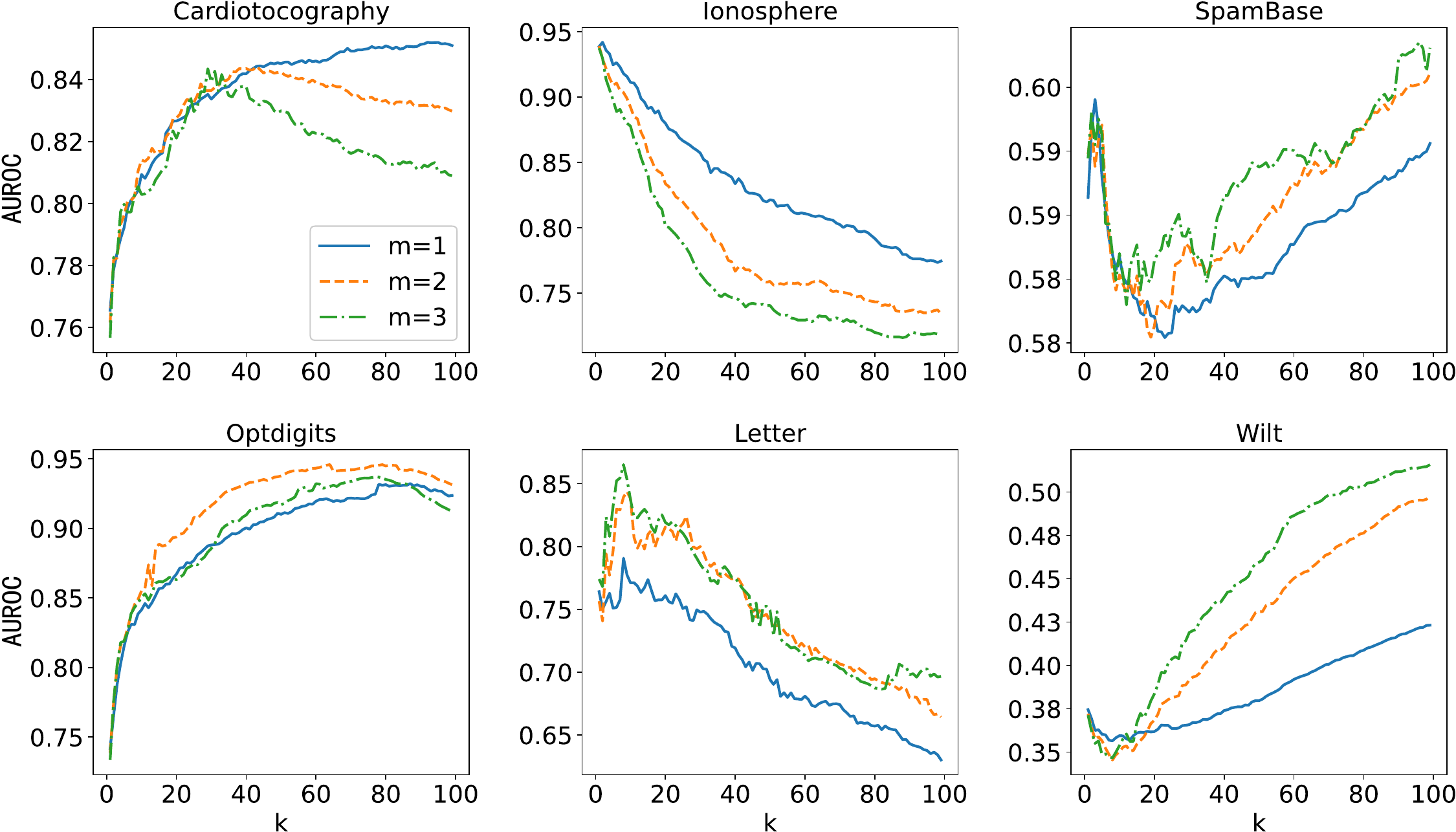}}
\caption{The performance curves as $k$ was increased from 1 to 99 on 6 datasets (\textit{Cardiotocography}, \textit{Ionosphere}, \textit{SpameBase}, \textit{Optdigits}, \textit{Letter} and \textit{ Wilt}). Each individual graph has three curves with different $m$ denoting the time of mean-shifting. For each dataset, the $k$-NN relationships were computed on the corresponding training set ($\frac{3}{4}$ size of the complete set) and the AUROC results were evaluated on the corresponding validation set.}
\label{fig:k}
\end{figure}

\subsubsection{\textcolor{black}{Hyper-parameters}}
The proposed methods required three hyper-parameters, which were $\alpha$, $m$, and $k$.  For WNLL, five different $\alpha$ were tested: \{0.20, 0.33, 0.50, 0.66, 0.8\}. The results are shown in Section \ref{sssec:ep}. For MSS, the mean-shift time $m$ was varied from 1 to 3. The number of nearest neighbors $k$ was selected from a candidate list $\mathbf{k}$. In our experiments, $\mathbf{k}=[1,2,\cdots,\min(99, N-1)]$, where $N$ denotes the number of the instances in the training set. The performance was reported in Section \ref{sssec:ems}.

\subsubsection{\textcolor{black}{Training strategy}}
All baseline methods and the proposed methods were trained on the training set, and tuned with the aid of the validation set. After training, the best models were tested on the testing set to obtain the final results. In addition, to increase the stability of AE-based methods, we applied an ensemble technique. 20 initial states were generated for AEs and trained separately. After training, these models were validated by the validation set to find the best 5 models. Finally, 5 outlier scores were computed on the 5 models for one test sample, then they were normalized and averaged as the final score. The source codes of the proposed method are available on Github for reproducibility \footnote{github.com/Ra1demmmm/Autoencoder-based-Outlier-Detection}.

\subsection{Experimental results}
\label{ssec:res}

\subsubsection{Evaluation of the probabilistic autoencoder}
\label{sssec:ep}
In this section, the performance of PAE was evaluated. As shown in TABLE \ref{tab:pae}, the average performance over 32 real-world datasets of PAE with all five values of $\alpha$ was better than the general AE. The largest relative improvement of the average AUROC was 29\% (0.851 vs. 0.791), comparing PAE ($\alpha=0.20$) and AE. Moreover, PAE significantly outperformed AE on some datasets, such as \textit{Annothyroid} (0.850 vs. 0.608), \textit{Glass} (0.980 vs. 0.804), \textit{Optdigits} (0.936 vs. 0.760), \textit{SpamBase} (0.847 vs. 0.565) and \textit{Wilt} (0.879 vs. 0.383), which showed its superiority for specific applications. 

The optimal value of $\alpha$ was application-dependent. For example, on dataset \textit{ALOI}, \textit{Cardiotocography}, \textit{Wilt} and \textit{Wine}, the performance improved as $\alpha$ increased, which implied that the reconstruction error or the negative effect of aleatoric uncertainty were more critical in these applications. In contrast, for datasets on dataset \textit{Optdigits}, \textit{PenDigits}, \textit{Satellite} and \textit{Waveform}, the performance improved as $\alpha$ decreased, which implied that the positive effect of aleatoric uncertainty was more critical in these applications. The logical deduction was that a suitable value of $\alpha$ should be selected based on the specific application to achieve the best OD performance.

The experiments illustrated that the average performance of all the other groups was better than the value of $\alpha=0.50$ (which was equivalent to NLL), which proved the importance of the adjustable parameters in WNLL. In addition, the value of $\alpha=0.20$ achieved the best performance on 13 datasets (41\%) and the best average AUROC performance (0.851). This clearly demonstrated that the uncertainty of AE was critical for OD, and even more important than the commonly used bias measurement in many applications. Insufficient emphasis had been placed on this variable, and significant change could be brought about by the discovery in this experiment. This finding has potential to facilitate all AE-based methods in OD applications.


\begin{table}[t]
	\centering
	\caption{AUROC results of AE vs. MSS-AE and PAE vs. MSS-PAE.}
	\label{tab:ms}
	\scalebox{0.7}{
		\begin{tabular}{c|c|ccc|l|c|ccc}
			\toprule
			\multirow{2}{*}{\textbf{Dataset}} & \multirow{2}{*}{\textbf{AE}} & \multicolumn{3}{c|}{\textbf{MSS-AE}} &  & \multirow{2}{*}{\textbf{PAE}} & \multicolumn{3}{c}{\textbf{MSS-PAE}} \\ \cline{3-5} \cline{8-10} 
			&  & $m=1$ & $m=2$ & $m=3$ &  &  & $m=1$ & $m=2$ & $m=3$ \\ \cmidrule{1-5} \cmidrule{7-10} 
			ALOI & \textbf{0.563} & 0.558 & 0.563 & 0.561 &  & 0.545 & 0.630 & 0.658 & \textbf{0.664} \\
			Annthyroid & 0.608 & 0.603 & 0.630 & \textbf{0.641} &  & 0.799 & 0.856 & 0.886 & \textbf{0.893} \\
			Arrhythmia & 0.714 & 0.720 & 0.724 & \textbf{0.730} &  & 0.736 & 0.716 & 0.761 & \textbf{0.762} \\
			Breastw & 0.972 & \textbf{0.994} & 0.987 & 0.987 &  & 0.994 & 0.994 & \textbf{0.996} & 0.992 \\
			Cardiotocography & 0.811 & 0.875 & \textbf{0.888} & 0.877 &  & 0.818 & 0.841 & \textbf{0.844} & \textbf{0.844} \\
			Glass & 0.804 & 0.922 & 0.863 & \textbf{0.951} &  & \textbf{0.961} & 0.804 & 0.824 & 0.843 \\
			HeartDisease & 0.813 & 0.859 & \textbf{0.869} & 0.799 &  & 0.865 & \textbf{0.892} & 0.883 & 0.886 \\
			InternetAds & 0.706 & 0.724 & \textbf{0.801} & 0.783 &  & 0.685 & \textbf{0.797} & \textbf{0.797} & \textbf{0.797} \\
			Ionosphere & 0.971 & \textbf{0.978} & 0.975 & 0.973 &  & \textbf{0.985} & 0.962 & 0.964 & 0.962 \\
			Letter & 0.857 & 0.844 & \textbf{0.860} & 0.853 &  & \textbf{0.855} & 0.748 & 0.742 & 0.798 \\
			Lymphography & 0.971 & \textbf{1.000} & \textbf{1.000} & \textbf{1.000} &  & 0.914 & 0.971 & \textbf{1.000} & \textbf{1.000} \\
			Mammography & \textbf{0.893} & 0.876 & 0.885 & 0.885 &  & 0.884 & 0.882 & \textbf{0.885} & 0.882 \\
			Mnist & \textbf{0.882} & 0.876 & 0.881 & \textbf{0.882} &  & 0.948 & \textbf{0.949} & \textbf{0.949} & 0.948 \\
			Musk & \textbf{1.000} & \textbf{1.000} & \textbf{1.000} & \textbf{1.000} &  & \textbf{1.000} & \textbf{1.000} & \textbf{1.000} & \textbf{1.000} \\
			Optdigits & 0.760 & 0.957 & 0.970 & \textbf{0.981} &  & 0.936 & \textbf{0.994} & 0.993 & 0.992 \\
			PageBlocks & 0.952 & \textbf{0.955} & 0.948 & 0.953 &  & \textbf{0.962} & 0.958 & 0.957 & 0.954 \\
			Parkinson & 0.627 & 0.639 & \textbf{0.769} & 0.736 &  & 0.815 & 0.766 & 0.796 & \textbf{0.819} \\
			PenDigits & 0.879 & 0.887 & \textbf{0.974} & 0.973 &  & 0.951 & 0.963 & \textbf{0.977} & 0.971 \\
			Pima & 0.651 & 0.689 & 0.691 & \textbf{0.699} &  & 0.756 & \textbf{0.757} & 0.740 & 0.722 \\
			Satellite & 0.750 & 0.788 & \textbf{0.789} & 0.763 &  & 0.855 & 0.856 & \textbf{0.857} & \textbf{0.857} \\
			Satimage-2 & \textbf{0.993} & \textbf{0.993} & 0.991 & 0.989 &  & \textbf{0.983} & 0.981 & 0.975 & 0.967 \\
			Shuttle & 0.995 & 0.997 & \textbf{1.000} & 0.997 &  & 0.996 & \textbf{1.000} & 0.997 & 0.997 \\
			SpamBase & 0.565 & 0.580 & 0.622 & \textbf{0.623} &  & 0.847 & \textbf{0.886} & 0.879 & 0.878 \\
			Speech & 0.551 & 0.567 & 0.586 & \textbf{0.651} &  & 0.541 & \textbf{0.658} & 0.477 & 0.527 \\
			Stamps & 0.931 & \textbf{0.961} & 0.915 & 0.950 &  & 0.844 & \textbf{0.954} & 0.942 & 0.837 \\
			Thyroid & \textbf{0.975} & 0.972 & 0.963 & 0.960 &  & 0.973 & \textbf{0.994} & 0.991 & 0.990 \\
			Vertebral & 0.607 & \textbf{0.755} & 0.750 & 0.745 &  & 0.676 & 0.687 & \textbf{0.712} & 0.703 \\
			Vowels & 0.937 & 0.906 & 0.925 & \textbf{0.958} &  & \textbf{0.968} & 0.945 & 0.940 & 0.943 \\
			Waveform & 0.781 & 0.792 & 0.826 & \textbf{0.872} &  & 0.881 & 0.895 & 0.897 & \textbf{0.901} \\
			Wilt & 0.383 & 0.427 & 0.478 & \textbf{0.511} &  & 0.822 & \textbf{0.884} & 0.873 & 0.676 \\
			Wine & 0.914 & 0.948 & \textbf{1.000} & \textbf{1.000} &  & 0.879 & 0.948 & \textbf{0.966} & \textbf{0.966} \\
			WPBC & 0.499 & \textbf{0.506} & 0.413 & 0.477 &  & \textbf{0.548} & 0.415 & 0.428 & 0.450 \\ \cmidrule{1-5} \cmidrule{7-10} 
			AVG. & 0.791 & 0.817 & 0.829 & \textbf{0.836} &  & 0.851 & \textbf{0.862} & \textbf{0.862} & 0.857 \\ 
			BEST & 6 & 9 & 11 & \textbf{14} &  & 8 & \textbf{12} & 11 & 11 \\ 
			\bottomrule
		\end{tabular}
	}
\end{table}

\subsubsection{Evaluation of the mean-shift outlier scoring}
\label{sssec:ems}
The number of nearest neighbors $k$ and the time of mean-shifting $m$ were two important parameters for the MSS method. In reality, they were all application-dependent. The effect of $k$ was demonstrated for MSS-AE on several datasets in Fig. \ref{fig:k}. The figure illustrated the variation in performance as $k$ was increased for different datasets, and the distinct optimal $k$ for each dataset. Similar to all the $k$-NN-based methods, this factor depended on the data structure of the specific application.


The AUROC results for MSS-AE and MSS-PAE ($\alpha=0.20$) were listed in TABLE \ref{tab:ms}. Variation of the time of mean-shifting $m=1,2,3$ was tested for all three methods, and the optimal $k$ for each case was found using the validation set. The average results on 32 datasets demonstrated that MSS-AE and MSS-PAE triumphed their counterparts AE and PAE respectively. The best average AUROC performance for MSS-AE was 0.836 when $m=3$, and for MSS-PAE, it was 0.858 when $m=1$, which was the best among all competitors. For a total of 32 datasets, MSS-AE was better than AE for 27 datasets (84\%), and MSS-PAE was better than PAE for 24 datasets (75\%), which proved the effectiveness of the MSS. TABLE \ref{tab:ms} illustrated that the MSS significantly improved the performance on the datasets where AE or PAE had poor performance, such as \textit{InternetAds}, \textit{Optidigits}, \textit{Speech} and \textit{Vertebral}. This proved that the combination of the AE-based method and the MSS was effective.

\begin{table}[t]
	\centering
	\caption{AUROC results of 6 AE-based methods and their MSS- versions.}
	\label{tab:aes}
	\resizebox{\textwidth}{!}{
		\begin{tabular}{c|cc|cc|cc|cc|cc|cc}
			\toprule
			\multirow{2}{*}{\textbf{Dataset}} & \textbf{AE}    & \textbf{MSS-AE} & \textbf{PAE}   & \textbf{MSS-PAE} & \textbf{RDA}   & \textbf{MSS-RDA} & \textbf{LCAE}  & \textbf{MSS-LCAE} & \textbf{RSRAE} & \textbf{MSS-RSRAE} & \textbf{FDAE}  & \textbf{MSS-FDAE} \\
			& \cite{hawkins2002outlier}              & *               & *               & *                & \cite{zhou2017anomaly}              & *                & \cite{ishii2019low}              & *                 & \cite{yu2021autoencoder}              & *                  & \cite{guo2022unsupervised}              &  *                 \\ \midrule
			ALOI             & 0.563          & 0.563          & 0.564          & \textbf{0.706}  & 0.562          & 0.557           & 0.553          & 0.584            & 0.549          & 0.571             & 0.547          & 0.540            \\
			Annthyroid       & 0.608          & 0.641          & 0.850          & \textbf{0.893}  & 0.608          & 0.617           & 0.749          & 0.689            & 0.611          & 0.606             & 0.623          & 0.617            \\
			Arrhythmia       & 0.714          & 0.730          & 0.740          & \textbf{0.762}  & 0.714          & 0.729           & 0.724          & 0.713            & 0.711          & 0.724             & 0.709          & 0.639            \\
			Breastw          & 0.972          & 0.994          & 0.995          & \textbf{0.996}  & 0.972          & 0.995           & 0.988          & 0.993            & 0.995          & \textbf{0.996}    & 0.993          & 0.989            \\
			Cardiotocography & 0.811          & 0.888          & 0.870          & 0.899           & 0.811          & 0.888           & 0.812          & 0.883            & 0.812          & \textbf{0.901}    & 0.812          & \textbf{0.901}   \\
			Glass            & 0.804          & 0.951          & \textbf{0.980} & 0.853           & 0.863          & \textbf{0.980}  & 0.922          & 0.863            & 0.745          & 0.941             & 0.755          & 0.951            \\
			HeartDisease     & 0.813          & 0.869          & 0.871          & 0.905           & 0.813          & 0.885           & 0.770          & \textbf{0.909}   & 0.799          & 0.845             & 0.750          & 0.848            \\
			InternetAds      & 0.706          & 0.801          & 0.697          & \textbf{0.839}  & 0.707          & 0.801           & 0.702          & 0.765            & 0.653          & 0.754             & 0.665          & 0.711            \\
			Ionosphere       & 0.971          & 0.978          & \textbf{0.985} & 0.978           & 0.968          & 0.968           & 0.980          & 0.977            & 0.965          & 0.962             & 0.949          & 0.952            \\
			Letter           & 0.857          & 0.860          & 0.855          & 0.798           & 0.820          & 0.833           & 0.855          & 0.832            & \textbf{0.867} & 0.794             & 0.829          & 0.793            \\
			Lymphography     & 0.971          & \textbf{1.000} & 0.914          & \textbf{1.000}  & 0.971          & \textbf{1.000}  & 0.971          & \textbf{1.000}   & \textbf{1.000} & \textbf{1.000}    & \textbf{1.000} & \textbf{1.000}   \\
			Mammography      & 0.893          & 0.885          & 0.891          & 0.901           & 0.892          & 0.890           & 0.896          & 0.895            & \textbf{0.907} & \textbf{0.907}    & 0.890          & 0.890            \\
			Mnist            & 0.882          & 0.882          & 0.948          & \textbf{0.949}  & 0.882          & 0.889           & 0.925          & 0.919            & 0.891          & 0.886             & 0.892          & 0.882            \\
			Musk             & \textbf{1.000} & \textbf{1.000} & \textbf{1.000} & \textbf{1.000}  & \textbf{1.000} & \textbf{1.000}  & \textbf{1.000} & \textbf{1.000}   & \textbf{1.000} & \textbf{1.000}    & \textbf{1.000} & \textbf{1.000}   \\
			Optdigits        & 0.760          & 0.981          & 0.936          & 0.995           & 0.760          & 0.981           & 0.759          & 0.950            & 0.795          & \textbf{0.998}    & 0.818          & 0.990            \\
			PageBlocks       & 0.952          & 0.955          & 0.962          & 0.958           & 0.952          & 0.956           & 0.944          & 0.948            & 0.959          & \textbf{0.961}    & 0.958          & \textbf{0.961}   \\
			Parkinson        & 0.627          & 0.769          & 0.815          & \textbf{0.861}  & 0.620          & 0.769           & 0.755          & 0.725            & 0.690          & 0.778             & 0.653          & 0.764            \\
			PenDigits        & 0.879          & 0.974          & 0.951          & \textbf{0.993}  & 0.841          & 0.970           & 0.925          & 0.976            & 0.764          & 0.946             & 0.831          & 0.960            \\
			Pima             & 0.651          & 0.699          & 0.756          & \textbf{0.757}  & 0.651          & 0.693           & 0.677          & 0.677            & 0.681          & 0.723             & 0.661          & 0.667            \\
			Satellite        & 0.750          & 0.789          & 0.855          & 0.857           & 0.745          & 0.791           & 0.830          & \textbf{0.858}   & 0.817          & 0.829             & 0.840          & 0.847            \\
			Satimage-2       & 0.993          & 0.993          & 0.988          & 0.987           & 0.993          & 0.993           & \textbf{0.994} & 0.993            & 0.992          & 0.990             & 0.993          & 0.978            \\
			Shuttle          & 0.995          & \textbf{1.000} & 0.996          & \textbf{1.000}  & 0.995          & 0.997           & 0.960          & \textbf{1.000}   & 0.989          & \textbf{1.000}    & 0.977          & 0.987            \\
			SpamBase         & 0.565          & 0.623          & 0.847          & \textbf{0.886}  & 0.554          & 0.623           & 0.548          & 0.734            & 0.594          & 0.771             & 0.570          & 0.716            \\
			Speech           & 0.551          & 0.651          & 0.597          & \textbf{0.673}  & 0.525          & 0.651           & 0.529          & 0.634            & 0.507          & 0.528             & 0.563          & 0.488            \\
			Stamps           & 0.931          & 0.961          & 0.952          & 0.954           & 0.957          & 0.961           & 0.941          & \textbf{0.974}   & 0.941          & 0.957             & 0.941          & 0.918            \\
			Thyroid          & 0.975          & 0.972          & 0.991          & \textbf{0.994}  & 0.976          & 0.972           & 0.980          & 0.980            & 0.982          & 0.982             & 0.971          & 0.970            \\
			Vertebral        & 0.607          & 0.755          & 0.772          & \textbf{0.841}  & 0.593          & 0.755           & 0.676          & 0.808            & 0.602          & 0.745             & 0.703          & 0.695            \\
			Vowels           & 0.937          & 0.958          & \textbf{0.976} & 0.945           & 0.925          & 0.939           & 0.919          & 0.929            & 0.901          & 0.886             & 0.727          & 0.707            \\
			Waveform         & 0.781          & 0.872          & 0.881          & \textbf{0.901}  & 0.781          & 0.817           & 0.834          & 0.758            & 0.783          & 0.859             & 0.802          & 0.812            \\
			Wilt             & 0.383          & 0.511          & 0.879          & \textbf{0.891}  & 0.388          & 0.503           & 0.483          & 0.559            & 0.529          & 0.529             & 0.529          & 0.536            \\
			Wine             & 0.914          & \textbf{1.000} & 0.914          & \textbf{1.000}  & 0.914          & \textbf{1.000}  & 0.897          & 0.983            & 0.914          & \textbf{1.000}    & 0.966          & \textbf{1.000}   \\
			WPBC             & 0.499          & 0.506          & 0.548          & 0.484           & 0.499          & 0.506           & 0.526          & 0.575            & 0.612          & 0.521             & 0.585          & \textbf{0.764}   \\ \midrule
			AVG.             & 0.791          & 0.844          & 0.868          & \textbf{0.889}  & 0.789          & 0.841           & 0.813          & 0.846            & 0.799          & 0.840             & 0.797          & 0.827            \\ 
			BEST             & 1          & 4          & 4          & \textbf{19}  & 1          & 4           & 2          & 6            & 4          & 9             & 2          & 6            \\ 
			\bottomrule
		\end{tabular}
	}
	
\end{table}

\begin{table}[!h]
	\centering
	\caption{AUROC results of the proposed methods and non-AE-based baselines.}
	\label{tab:sts}
	\resizebox{\textwidth}{!}{
		\begin{tabular}{c|ccccccccp{0.9cm}<{\centering}|ccp{1.7cm}<{\centering}}
			\toprule
			\textbf{Dataset} & {\begin{tabular}[c]{@{}c@{}}\textbf{ECOD}\\ \cite{li2022ecod}\end{tabular}} & {\begin{tabular}[c]{@{}c@{}}\textbf{iForest}\\ \cite{liu2008isolation}\end{tabular}} & {\begin{tabular}[c]{@{}c@{}}\textbf{LOF}\\ \cite{breunig2000lof}\end{tabular}} & {\begin{tabular}[c]{@{}c@{}}\textbf{OCSVM}\\ \cite{scholkopf2001estimating}\end{tabular}} & {\begin{tabular}[c]{@{}c@{}}\textbf{HBOS}\\ \cite{goldstein2012histogram}\end{tabular}} & {\begin{tabular}[c]{@{}c@{}}\textbf{MOD}\\ \cite{yang2021mean}\end{tabular}} & {\begin{tabular}[c]{@{}c@{}}\textbf{DIF}\\ \cite{xu2023deep}\end{tabular}} & {\begin{tabular}[c]{@{}c@{}}\textbf{DIP}\\ \cite{zhao2023searching}\end{tabular}} & {\begin{tabular}[c]{@{}c@{}}\textbf{AE}\\ \cite{hawkins2002outlier}\end{tabular}} & {\begin{tabular}[c]{@{}c@{}}\textbf{PAE}\\ *\end{tabular}} & {\begin{tabular}[c]{@{}c@{}}\textbf{MSS-AE}\\ *\end{tabular}} & {\begin{tabular}[c]{@{}c@{}}\textbf{MSS-PAE}\\ *\end{tabular}} \\ \midrule
			ALOI & 0.540 & 0.544 & 0.785 & 0.543 & 0.494 & 0.769 & 0.546 & \textbf{0.788} & 0.563 & 0.564 & 0.563 & 0.706 \\
			Annthyroid & 0.739 & 0.652 & 0.696 & 0.601 & 0.670 & 0.716 & 0.504 & 0.728 & 0.608 & 0.850 & 0.641 & \textbf{0.893} \\
			Arrhythmia & 0.732 & 0.740 & 0.705 & 0.711 & 0.755 & 0.718 & 0.530 & 0.731 & 0.714 & 0.740 & 0.730 & \textbf{0.762} \\
			Breastw & 0.992 & 0.989 & 0.696 & 0.995 & 0.983 & 0.987 & 0.975 & 0.993 & 0.972 & 0.995 & 0.994 & \textbf{0.996} \\
			Cardiotocography & 0.802 & 0.706 & 0.628 & 0.767 & 0.619 & 0.598 & 0.689 & 0.556 & 0.811 & 0.870 & 0.888 & \textbf{0.899} \\
			Glass & 0.853 & 0.922 & 0.706 & 0.863 & 0.848 & 0.941 & 0.931 & 0.814 & 0.804 & \textbf{0.980} & 0.951 & 0.853 \\
			HeartDisease & 0.687 & 0.721 & 0.777 & 0.730 & 0.774 & 0.758 & 0.723 & 0.845 & 0.813 & 0.871 & 0.869 & \textbf{0.905} \\
			InternetAds & 0.722 & 0.764 & 0.632 & 0.653 & 0.736 & 0.660 & 0.750 & 0.696 & 0.706 & 0.697 & 0.801 & \textbf{0.839} \\
			Ionosphere & 0.789 & 0.895 & 0.931 & 0.885 & 0.892 & 0.932 & 0.903 & 0.906 & 0.971 & \textbf{0.985} & 0.978 & 0.978 \\
			Letter & 0.549 & 0.575 & 0.878 & 0.571 & 0.555 & \textbf{0.905} & 0.554 & 0.872 & 0.857 & 0.855 & 0.860 & 0.798 \\
			Lymphography & \textbf{1.000} & \textbf{1.000} & \textbf{1.000} & \textbf{1.000} & \textbf{1.000} & \textbf{1.000} & 0.914 & \textbf{1.000} & 0.971 & 0.914 & \textbf{1.000} & \textbf{1.000} \\
			Mammography & \textbf{0.914} & 0.864 & 0.805 & 0.863 & 0.811 & 0.844 & 0.776 & 0.843 & 0.893 & 0.891 & 0.885 & 0.901 \\
			Mnist & 0.751 & 0.819 & 0.846 & 0.857 & 0.534 & 0.860 & 0.584 & 0.851 & 0.882 & 0.948 & 0.882 & \textbf{0.949} \\
			Musk & 0.967 & 1.000 & 0.997 & \textbf{1.000} & \textbf{1.000} & 0.997 & 0.966 & \textbf{1.000} & \textbf{1.000} & \textbf{1.000} & \textbf{1.000} & \textbf{1.000} \\
			Optdigits & 0.615 & 0.799 & 0.542 & 0.557 & 0.915 & 0.494 & 0.649 & 0.835 & 0.760 & 0.936 & 0.981 & \textbf{0.995} \\
			PageBlocks & 0.914 & 0.905 & 0.933 & 0.930 & 0.766 & 0.921 & 0.929 & 0.915 & 0.952 & \textbf{0.962} & 0.955 & 0.958 \\
			Parkinson & 0.350 & 0.414 & 0.789 & 0.384 & 0.676 & 0.669 & 0.484 & 0.609 & 0.627 & 0.815 & 0.769 & \textbf{0.861} \\
			PenDigits & 0.392 & 0.758 & 0.936 & 0.539 & 0.761 & 0.976 & 0.764 & 0.985 & 0.879 & 0.951 & 0.974 & \textbf{0.993} \\
			Pima & 0.537 & 0.592 & 0.641 & 0.631 & 0.656 & 0.634 & 0.637 & 0.690 & 0.651 & 0.756 & 0.699 & \textbf{0.757} \\
			Satellite & 0.608 & 0.724 & 0.566 & 0.694 & 0.806 & 0.695 & 0.748 & 0.696 & 0.750 & 0.855 & 0.789 & \textbf{0.857} \\
			Satimage-2 & 0.969 & 0.992 & 0.976 & 0.996 & 0.980 & 0.998 & 0.996 & \textbf{0.998} & 0.993 & 0.988 & 0.993 & 0.987 \\
			Shuttle & 0.723 & 0.820 & 0.993 & 0.989 & 0.803 & 0.989 & 0.900 & 0.988 & 0.995 & 0.996 & \textbf{1.000} & \textbf{1.000} \\
			SpamBase & 0.679 & 0.622 & 0.474 & 0.537 & 0.690 & 0.538 & 0.552 & 0.662 & 0.565 & 0.847 & 0.623 & \textbf{0.886} \\
			Speech & 0.507 & 0.527 & 0.698 & 0.506 & 0.518 & \textbf{0.734} & 0.582 & 0.627 & 0.551 & 0.597 & 0.651 & 0.673 \\
			Stamps & 0.870 & 0.931 & 0.944 & 0.941 & 0.786 & 0.937 & 0.917 & \textbf{0.961} & 0.931 & 0.952 & \textbf{0.961} & 0.954 \\
			Thyroid & 0.977 & 0.977 & 0.964 & 0.953 & 0.982 & 0.963 & 0.941 & 0.973 & 0.975 & 0.991 & 0.972 & \textbf{0.994} \\
			Vertebral & 0.596 & 0.552 & 0.646 & 0.629 & 0.549 & 0.536 & 0.371 & 0.591 & 0.607 & 0.772 & 0.755 & \textbf{0.841} \\
			Vowels & 0.632 & 0.767 & 0.939 & 0.792 & 0.682 & \textbf{0.984} & 0.760 & 0.979 & 0.937 & 0.976 & 0.958 & 0.945 \\
			Waveform & 0.674 & 0.759 & 0.746 & 0.776 & 0.753 & 0.734 & 0.648 & 0.800 & 0.781 & 0.881 & 0.872 & \textbf{0.901} \\
			Wilt & 0.350 & 0.437 & 0.690 & 0.295 & 0.323 & 0.667 & 0.478 & 0.640 & 0.383 & 0.879 & 0.511 & \textbf{0.891} \\
			Wine & 0.897 & 0.862 & 0.931 & 0.879 & 0.966 & 0.931 & 0.897 & 0.879 & 0.914 & 0.914 & \textbf{1.000} & \textbf{1.000} \\
			WPBC & 0.482 & 0.479 & 0.526 & 0.499 & 0.536 & 0.509 & 0.494 & 0.528 & 0.499 & \textbf{0.548} & 0.506 & 0.484 \\ \midrule
			AVG. & 0.713 & 0.753 & 0.782 & 0.736 & 0.744 & 0.800 & 0.722 & 0.812 & 0.791 & 0.868 & 0.844 & \textbf{0.889}\\
			BEST & 2& 1 & 1 & 2 & 2 & 4 & 0 & 5 &1 & 5 & 5 & \textbf{21}\\
			\bottomrule
		\end{tabular}
	}
	
\end{table}

\subsubsection{Comparison with baseline methods}
\label{sssec:ca}
To show the effectiveness of the proposed WNLL function and the versatility of the MSS method, the performance of PAE, 5 AE-based OD methods, and their mean-shifted versions were compared in TABLE \ref{tab:aes}. The parameters of all methods in this section were tuned to be optimal. The results of all MSS- versions were obtained with the optimal value of $m$, and the results of PAE and MSS-PAE were obtained using the best value of $\alpha$. The table conclusively showed that all MSS- versions outperformed their original counterpart with average 20\% of the relative performance improvement on the average AUROC, proving the versatility of the MSS method. Then among all standard versions, PAE achieved the best average AUROC performance, which gained a 29\% relative improvement comparing to the second best method LCAE (0.868 vs. 0.813). Among all MSS- versions, MSS-PAE also achieved the best average AUROC performance, which gained a 28\% relative improvement comparing to the second best method MSS-LCAE (0.889 vs. 0.846). This result suggested that the proposed WNLL was effective, especially when jointly incorporated with the MSS method.

To further demonstrate the superiority of the proposed methods, the performance comparison of the proposed method against 8 non-AE-based OD methods were summarized in TABLE \ref{tab:sts}. The results showed that the proposed PAE, MSS-AE, and MSS-PAE outperformed all the baselines on average AUROC, where MSS-PAE achieved the best performance on 21 datasets among all 32 datasets (66\%) and had the best average AUROC of 0.889 (41\% relative improvement comparing with DIP which was the best of 8 baselines). The results strongly proved the potency of combining the feature extraction (neural network), uncertainty estimation (probabilistic analysis), and local structure information (nearest-neighbor-graph). It indicated that the OD researchers should consider multiple instruments to design an outlier detector, to adapt various complex realistic scenarios.

\section{Conclusion}
\label{sec:con}
In this paper, the issue of unexpected reconstruction was mitigated by two novel strategies. First, the overconfident issue and the role of aleatoric uncertainty for AE-based OD were analyzed. The analysis supported usage of NLL to train AE, and WNLL for outlier scoring. Several recommendations were provided for application of NLL and WNLL for various OD scenarios. Second, the MSS method was proposed to reduce the false inliers judged by the AE-based OD methods, by introducing the information of the local relationship.

The experimental results on 32 real-world OD datasets showed that the proposed methods significantly improved AE's performance on OD. Specifically, WNLL was quite effective and flexible when adapting to different OD applications, and the proposed MSS method could be effectively applied to other AE-based OD methods. Furthermore, MSS-PAE, which was the combination of the proposed two methods, performed best among all baselines. Therefore, the experiments supported the use of the proposed WNLL and MSS simultaneously to get the best performance in practice. We believe that the MSS method can be used as a general plugin for most AE-based OD methods. 
In addition, NLL and WNLL may also be applied to improve the performance of other AE-based methods, and can be integrated with other loss functions, which were not evaluated in this paper but can be validated in the future research.

%
%
%
%
%
%

\small
\bibliographystyle{elsarticle-num}
\bibliography{refs}

\end{document}